\documentclass{article} % For LaTeX2e
\usepackage{iclr2027_conference,times}

\usepackage[utf8]{inputenc}
\usepackage[T1]{fontenc}

\usepackage{microtype}
\usepackage{graphicx}
\usepackage{subcaption}
\usepackage{booktabs}
\usepackage{multirow}
\usepackage{makecell}
\usepackage{xcolor}
\usepackage{xspace}
\usepackage{url}
\usepackage{threeparttable}

\usepackage{amsmath}
\usepackage{amssymb}
\usepackage{mathtools}
\usepackage{amsthm}
\usepackage{amsfonts}
\usepackage{nicefrac}
\usepackage{enumitem}

\usepackage{algorithm}
\usepackage{algorithmic}

\usepackage{hyperref}
\usepackage[capitalize,noabbrev]{cleveref}

\usepackage{wrapfig}

\usepackage[dvipsnames]{xcolor}

\newcommand{\method}{\textsc{SQuAT}\xspace}
\newcommand{\optionA}{Standard VQ-QAT\xspace}
\newcommand{\optionB}{Spherical VQ-QAT\xspace}
\newcommand{\optionAs}{Standard\xspace}
\newcommand{\optionBs}{Spherical\xspace}

\title{\method: Spherical Lattice Vector \\ Quantization-Aware Training}

\author{
\hspace{-0.7mm}Sebastian Loeschcke%
\thanks{Equal contribution.} \hspace{1.5pt}$^{,}$%
\thanks{Work done while doing an internship at Qualcomm AI Research.} \\
University of Copenhagen \\
\texttt{sbl@di.ku.dk} \\
\And
Ivan Koryakovskiy$^{1}$ \\
Qualcomm AI Research%
\thanks{Qualcomm AI Research is an initiative of Qualcomm Technologies, Inc.} \\
\texttt{ivankor@qti.qualcomm.com\hphantom{ }} \\
\And
Tycho F.~A.~van~der~Ouderaa \\
Qualcomm AI Research$^3$ \\
\texttt{touderaa@qti.qualcomm.com} \\
\And
Riccardo Del~Chiaro \\
Qualcomm AI Research$^3$ \\
\texttt{rdelchia@qti.qualcomm.com} \\
\And
Mart van~Baalen \\
Qualcomm AI Research$^3$ \\
\texttt{mart@qti.qualcomm.com} \\
\And
Paul N.~Whatmough \\
Qualcomm AI Research$^3$ \\
\texttt{pwhatmou@qti.qualcomm.com} \\
\And
Markus Nagel \\
Qualcomm AI Research$^3$ \\
\texttt{markusn@qti.qualcomm.com} \\
}

\iclrfinalcopy % Uncomment for camera-ready version, but NOT for submission.
\arxivtrue
\begin{document}

\maketitle

\begin{abstract}
Vector quantization (VQ), in particular lattice-based methods, offer strong rate-distortion trade-offs for low-bit large language models (LLMs), yet their behavior under quantization-aware training (VQ-QAT) remains poorly understood. We scale lattice VQ-QAT to billion-parameter LLMs and multi-billion-token training regimes, and introduce \textbf{\method}, a shape-gain VQ-QAT approach trained with a separate straight-through estimator (STE) for the shape and gain. \method decomposes weights into direction and magnitude: normalized directions are quantized using an $E_8$-derived spherical codebook, while gains are quantized independently with low-bit scalar quantizers.
We show that, unlike a standard STE applied to the full quantizer, which ignores the shape-gain structure, these separate STEs induce tangent-space updates for directions, yielding an implicit form of spherical (Riemannian) optimization.
To enable scalability, we develop a distributed exact nearest-neighbor assignment scheme, in which workers quantize disjoint weight blocks and communicate only discrete indices, reducing VQ-QAT training time by 12.8$\times$ and allowing us to scale it to 10B tokens.
Across instruction-tuned LLaMA-3.2-1B/3B, Qwen3-4B, and base Qwen3-1.7B, \method consistently outperforms standard VQ-QAT STE under identical quantizers and initialization, and achieves SOTA VQ-QAT accuracy.
\end{abstract}

\section{Introduction}
Large language models (LLMs) contain billions of parameters~\citep{touvron2023llama,touvron2023llama2}, making weight storage and movement a dominant bottleneck for inference. 
Even with INT8 or INT4 formats, memory bandwidth remains a limitation for token generation, so aggressive quantization is critical to reduce footprint, while preserving quality~\citep{frantar2023gptqaccurateposttrainingquantization,lin2024awq,pmlr-v202-xiao23c}.

Most practical LLM quantization pipelines rely on integer scalar quantization.
However, recent methods have demonstrated that vector quantization (VQ) can offer superior quality--compression trade-offs by quantizing groups of weights jointly~\citep{zhang2025learning,vanbaalen2025gptvqblessingdimensionalityllm,aqlm}, especially at very low bit-widths.
While general VQ requires codebooks that grow exponentially with quantization dimensionality, \emph{Structured} VQ methods based on lattices~\citep{tseng2024quip-sharp-betterllmquantization}, trellises~\citep{tseng2025qtipquantizationtrellisesincoherence}, or pyramid constructions~\citep{vanderouderaa2024pyramidvectorquantizationllms} avoid storing explicit codebooks.
Some methods \citep{vanderouderaa2024pyramidvectorquantizationllms,vanderouderaa2026leechlatticevectorquantization} decompose weight vectors into \emph{shape} (direction) and \emph{gain} (magnitude), motivated by the observation that distortion is often dominated by angular error, while block norms can be represented with fewer bits~\citep{sabin1982product,yue2025pcdvqenhancingvectorquantization}.

\begin{figure*}[t]
  \centering
  \includegraphics[width=0.99\textwidth]{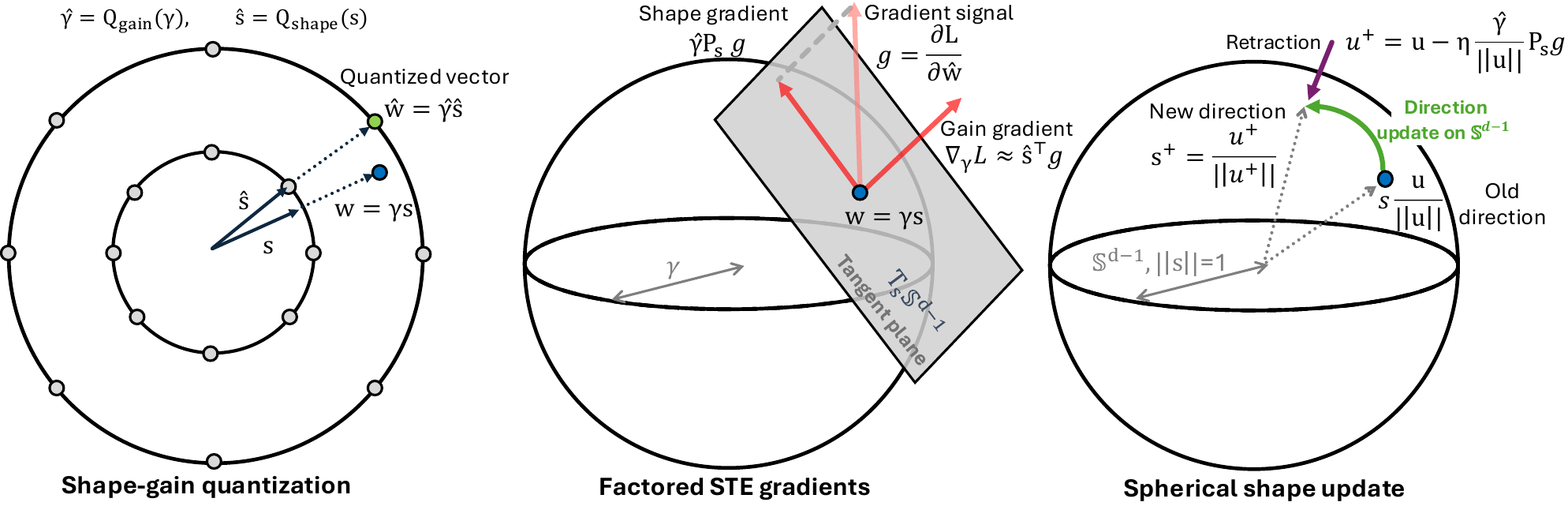}
  \vspace{-0.1cm}
  \caption{
\textbf{Overview of \method.}
Left: a weight block is decomposed as $w=\gamma s$, where $s\in\mathbb{S}^{d-1}$ is the shape and $\gamma \in \mathbb{R}^{+}$ is the gain. The two components are quantized independently and reconstructed as $\hat w=\hat\gamma\hat s$.
Middle: our spherical factored STE applies separate STEs to the gain and shape quantizers, mapping the gradient $g=\partial\mathcal L/\partial\hat w$ to a scalar gain gradient $\nabla_\gamma\mathcal L\approx \hat s^\top g$ and a tangent shape signal $\hat\gamma P_s g$ on $T_s\mathbb{S}^{d-1}$.
Right: the tangent shape signal induces the update
$u^+=u-\eta\,\hat\gamma(\|u\|)^{-1}P_s g$
on the shape parameter, followed by normalization (retraction back to the unit sphere), $s^+=u^+/(\|u^+\|)$.}
\label{fig:method_overview}
\vspace{-2.0em}
\end{figure*}

Quantization-aware training (QAT) for LLMs with carefully chosen low-bit quantizers and sufficient training budget can retain strong model quality~\citep{panferov2025quest,liu2025paretoq}. 
While some work on QAT for VQ exists (e.g., \citet{wang2026lcqat,pvtuning}), VQ-QAT remains largely unexplored in billion-parameter and multi-billion token regimes. Furthermore, existing work relies on multi-dimensional application of scalar QAT gradient estimation and parameter updates, with no prior work exploiting the shape-gain decomposition in a QAT setting. As a result, it remains unclear how to derive a QAT rule that aligns gradient estimation and parameter updates with spherical VQ geometry.

\begin{wrapfigure}{r}{0.40\columnwidth}
% \vspace{-3.5em}
\centering
\includegraphics[width=0.40\columnwidth]{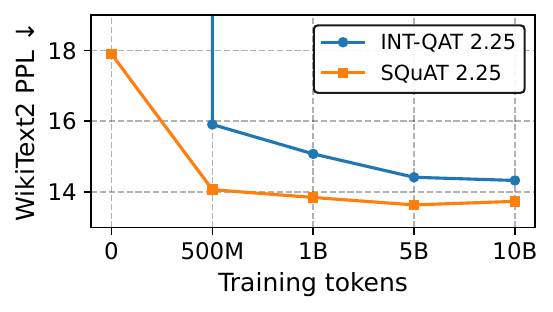}
% \vspace{-2.0em}
\caption{
\method improves WikiText2 perplexity over integer scalar (INT-QAT) at 2.25~bpw on LLaMA-3.2-1B.}
% \vspace{-0.5em}
\label{fig:main_result}
% \vspace{-1em}
\end{wrapfigure}

To address this, we introduce \textbf{\method}, a \textbf{S}pherical Lattice Vector \textbf{Qu}antization-\textbf{A}ware \textbf{T}raining
method for billion-parameter LLMs, that trains shape-gain quantizers with a \emph{factored} straight-through estimator (STE), illustrated in Figure~\ref{fig:method_overview}.
\method represents each weight block using separate trainable direction and gain parameters.
In the forward pass, the direction parameter is normalized and quantized with an $E_8$-derived spherical codebook, including normalized spherical E8P12 codebook~\citep{tseng2024quip-sharp-betterllmquantization} or our proposed E8S construction from equal-norm lattice shells, while the gain is quantized independently with a low-bit scalar quantizer.
In the backward pass, the STE is applied separately through the direction and gain quantizers, rather than jointly through the reconstructed weight.
This factored STE induces tangent-space updates for the direction parameter, yielding spherical optimization of quantized weight directions while decoupling magnitude optimization into the gain parameter.
While spherical parameterizations themselves are well established in machine learning (e.g., DoRA~\citep{liu2024doraweightdecomposedlowrankadaptation}) and quantization, prior work has not analyzed how combining such parameterizations with STE-based quantization alters the surrogate gradients and the resulting optimization dynamics.

To allow practical scaling to billion-parameter models and multi-billion token training,  we introduce distributed exact assignment, where workers quantize disjoint weight blocks against a shared codebook and communicate only discrete indices.

We evaluate \method on instruction-tuned LLaMA-3.2-1B/3B, Qwen3-4B, and base Qwen3-1.7B models, with QAT budgets from 0.25B to 10B tokens.
\method improves over standard joint-STE VQ-QAT under the same quantizer and initialization, substantially improves over VQ-PTQ baselines after QAT, and remains competitive with mature scalar INT-QAT at comparable bit budgets, see Figure~\ref{fig:main_result}.

\method makes the following contributions:\vspace{-0.3cm}
\begin{itemize}[leftmargin=1.5em, itemsep=0.5em, topsep=0.5em, parsep=0em, partopsep=0em]
    \item We propose a novel training rule for shape-gain lattice VQ-QAT, and show that this approach induces tangent-space updates on directions, yielding true \optionB.
    \item We propose a novel $E_8$-derived normalized spherical codebook which enables fractional bitwidth quantization, unlocking Pareto-optimal bit rates inaccessible to scalar methods.
    % \item We compress billion-parameter LLMs by combining \optionB with low-bit gain quantization and $E_8$-derived spherical codebooks.
    % \item We introduce a fast exact VQ-QAT implementation with fused kernels, index caching, and distributed assignment which yields a 12.8$\times$ speedup, allowing VQ-QAT to scale to billion-parameter models and multi-billion token training.
    \item We introduce a fast exact nearest-neighbor assignment implementation with fused kernels, index caching, and distributed assignment, yielding a 12.8$\times$ speedup and enabling exact codebook-based VQ-QAT at billion-parameter and multi-billion-token scale.
    \item Empirically, \method improves over the \optionA baseline under matched quantizers and initialization, and achieves SOTA VQ-QAT accuracy.
    % \item At 2.25\,bpw, \method outperforms INT-QAT across all models and training budgets   on WikiText2 perplexity, and achieves competitive or better MMLU and CSR, with MMLU gains of up to 1.6\,\% even after 10B token training.
      \item At 2.25\,bpw, \method outperforms INT-QAT across all models and training budgets
  on WikiText2 perplexity, and achieves competitive or better MMLU and CSR, with MMLU gains of up to 7.9\,\% at 1B token training.
\end{itemize}

\section{Background}
% \vspace{-0.25cm}
% We review the quantization and shape-gain preliminaries used by \method.
\vspace{-0.25cm}
\subsection{Scalar quantization}
\vspace{-0.2cm}
Scalar quantization maps each weight independently to a low-precision discrete set, typically using uniform integer levels~\citep{jacob2017quantizationtrainingneuralnetworks}.
During QAT, quantization is applied in the forward pass and gradients are propagated with a straight-through estimator (STE).
As our scalar INT-QAT baseline, we use Stretched Elastic Quantization (SEQ)~\citep{liu2025paretoq,lee2025upq}, which shifts the signed quantization grid by half a step and improves stability in low-bit regimes.

\vspace{-0.2cm}
\subsection{Vector and lattice quantization}
\vspace{-0.2cm}
Vector quantization (VQ) groups parameters into blocks $w \in \mathbb{R}^d$ and maps each block to its nearest codeword from a codebook
$\mathcal{C}=\{c_1,\ldots,c_K\}\subset \mathbb{R}^d$:
\begin{equation}
j^\star = \arg\min_{j \in \{1,\ldots,K\}} \|w - c_j\|_2,
\qquad
Q(w)=c_{j^\star}.
\end{equation}
Here, $j^\star$ is the discrete codeword index and $Q(w)$ is the reconstructed quantized block.
The storage cost is $\log_2(K)/d$ bits per weight.
By operating in higher dimensions, VQ can exploit the joint parameter distribution and achieve better rate–distortion trade-offs than scalar quantization.

Explicit learned or unstructured codebooks become infeasible in higher dimensions since the number of required codewords (and the cost of nearest-neighbor search) grow exponentially in dimension $d$. Lattice quantization instead uses implicitly specified, highly structured codebooks with strong packing properties, retaining good rate--distortion behavior \citep{GIBSON1988259,SpherePackings} while enabling scalable assignment procedures. Yet, VQ-QAT demands repeated nearest-neighbor (NN) assignments during training, making efficient assignment a central challenge.

\vspace{-0.2cm}
\subsection{Shape-gain (polar) decomposition}
\label{sec:shape-gain}
\vspace{-0.2cm}

A common perspective in vector quantization is to decompose a vector into its \emph{shape} (direction) and \emph{gain} (magnitude) components, also known as shape-gain quantization \citep{sabin1982product}, which has also been successfully applied to vector quantization of LLMs~\citep{vanderouderaa2024pyramidvectorquantizationllms, yue2025pcdvqenhancingvectorquantization}.
For $w\in\mathbb{R}^d$, write
\begin{align}
w = \gamma \,s,\qquad \gamma=\|w\|_2 \in \mathbb{R}^{+}, \quad s=\frac{w}{\gamma} \in\mathbb{S}^{d-1},
\end{align}
where $\gamma$ and $s$ are the gain and shape, respectively. 
These can now be quantized independently as
% Consider a forward (inference) quantizer of the form
\begin{equation}
\hat w = \hat\gamma\,\hat s,\qquad
\hat\gamma = Q_{\text{gain}}(\gamma),\quad
\hat s = Q_{\text{shape}}(s;\mathcal{C}),
\label{eq:shape_gain_forward}
\end{equation}
where $Q_{\text{shape}}:\mathbb{S}^{d-1}\to\mathcal{C}$ maps each direction to a codeword in the discrete codebook $\mathcal{C}\subset\mathbb{S}^{d-1}$, with $\mathbb{S}^{d-1} = \left\{ x \in \mathbb{R}^d \mid \|x\|_2 = 1 \right\}$, and $Q_{\text{gain}}:\mathbb{R}_{> 0}\to\mathcal{C}_{\text{gain}}$ is a scalar quantizer mapping each gain to a quantized gain in $\mathcal{C}_{\text{gain}}\subset\mathbb{R}_{> 0}$.

This separates angular error from gain error: prior VQ work shows that most representational budget should be allocated to direction, while gains can be stored with few bits~\citep{hamkins2001optimal}. Shape-gain decompositions have also been useful for optimization and adaptation, e.g., weight normalization~\citep{salimans2016weight} and DoRA~\citep{liu2024doraweightdecomposedlowrankadaptation}. However, it remains unclear how to use this geometry in QAT, where weights are repeatedly updated through STEs. In this study, we analyze how STE updates commonly used in VQ interact with shape-gain.
 
\vspace{-0.2cm}
\subsection{Lattice codebooks from $E_8$}
\label{sec:e8-codebooks}
\vspace{-0.2cm}

The $E_8$ lattice is a highly symmetric lattice in $\mathbb{R}^8$ that achieves the densest known sphere packing in eight dimensions~\citep{SpherePackings}.
Its points are organized into norm shells, where $S_r=\{x\in E_8:\|x\|_2^2=r\}$.
The first nonzero shells contain $240$, $2160$, $6720$, and $17520$ points, corresponding to the coefficients of the $E_8$ theta series. Using a single shell, or a union of shells, as a codebook yields a spherical code where nearest-neighbor assignment depends on angular alignment.
Since $E_8$ is infinite, finite low-bit codebooks require truncation.
We use them either in Euclidean form, retaining lattice norms, or normalized form, projecting codewords to the unit sphere for shape-gain (SG) quantization.
In our notation, non-SG variants use Euclidean form, while SG variants use the normalized spherical form.

We study two $E_8$-derived constructions.
First, QuIP\#~\citep{tseng2024quip-sharp-betterllmquantization} introduces E8P12, a structured 2-bit-per-weight vector quantizer with $2^{16}$ possible codewords per 8D block.
E8P12 is derived from an $E_8$-related shifted $\hat D_8$ construction and is designed for compact decoding: its codewords are represented using a small base table together with parity, sign, and shift operations.
Second, we introduce E8S, a spherical $E_8$ truncation that enumerates points from low-norm $E_8$ shells, optionally subsamples the final shell to match fractional bit budgets, and normalizes the selected points to form a spherical code. See App.~\ref{sec:e8-codebook} for details.
\section{\method: Spherical lattice VQ-QAT}
\label{sec:method}

Normally, VQ-QAT maintains full-precision \emph{shadow} parameters in Euclidean space while emulating low-precision inference by applying quantization only in the forward pass. For vector quantizers, the forward map is typically discrete and thus non-differentiable; gradient-based training therefore requires an explicit surrogate for backpropagation. The most common choice is the straight-through estimator (STE). In shape-gain vector quantization, however, STE can be inserted at multiple plausible locations in the computation graph, and these choices lead to different surrogate gradients and, consequently, different optimization dynamics.

In this work,  we study how STE choices affect optimization for shape-gain vector quantization.
We analyze two natural cases: in Sec.~\ref{sec:optA} we consider `\optionA:' the standard joint STE applied to the overall vector quantizer, and in Sec.~\ref{sec:optB} we consider our `\optionB' in which we apply the STE separately to the shape and gain quantizers, coupled with a spherical parameterization of the weights. We show that these choices induce distinct surrogate gradients and therefore different \emph{effective} updates on the implied Euclidean weight vector. Moreover, we interpret the resulting update of \optionB as a hyperspherical (Riemannian) update: a tangent-space step for the direction combined with a radial update for the gain, where normalization acts as a retraction. We adopt this update as our training rule, denoted \method, and validate its benefit empirically. We emphasize that \method is a training rule agnostic to the specific spherical codebook used.%; in this paper we instantiate it with two $E_8$-derived codebooks, E8S and E8P12.

\vspace{-0.2cm}
\subsection{\optionA: Euclidean shadow weights and standard STE}
\label{sec:optA}

Let the shadow weights be partitioned into disjoint vectors $w\in\mathbb{R}^d$, where $d$ is the vector-quantizer dimensionality. Let
\begin{equation}
g \;\coloneqq\; \frac{\partial \mathcal{L}}{\partial \hat w}\in\mathbb{R}^d
\label{eq:upstream_grad}
\end{equation}
denote the gradient backpropagated to the quantized (dequantized) vector $\hat w = Q(w)$.

In the standard formulation, the Euclidean vectors $w$ serve as the shadow parameters, and the forward pass applies a vector quantizer $Q(\cdot)$ to produce $\hat w$. The straight-through estimator (STE) is applied to the non-differentiable mapping $Q(w)$, equivalently, to $\hat w$ viewed as a function of $w$, by approximating its Jacobian as the identity:
\begin{equation}
\frac{\partial Q(w)}{\partial w} \;\approx\; I
\quad\Longrightarrow\quad
\frac{\partial \mathcal{L}}{\partial w}
\approx g.
\label{eq:optA_grad}
\end{equation}
Thus, under SGD the shadow vector is updated directly by the upstream gradient.
We call this \optionA. Importantly, because the STE treats the entire mapping $Q(\cdot)$ as the identity in the backward pass, any internal structure of the quantizer is ignored during optimization. For example, even if we were to factor the forward mapping into a shape-gain form,
\begin{equation}
    Q(w)
= Q_{\text{shape}}\!\left(\frac{w}{\|w\|}\right)\,
  Q_{\text{gain}}\!\left(\|w\|\right),
\end{equation}
as long as the STE is applied to the overall mapping $Q(w)$, using a shape-gain quantizer in itself has no effect on the resulting gradient and effective update of $w$.

\vspace{-0.2cm}
\subsection{\optionB: spherical parameterization and STE}
\vspace{-0.2cm}
\label{sec:setup}
\label{sec:optB}

We note, however, that the optimization dynamics \textit{do} change once we either (i) parameterize shadow weights in shape-gain variables, or (ii) apply STE separately to the shape and gain quantizers.

\vspace{-0.3cm}
\paragraph{Spherical parameterization}
In this context, it is natural to also parameterize the shadow weights in terms of an (unnormalized) \emph{shape} $u$ and \emph{gain} $\gamma$ as shadow parameters:
\begin{equation}
u\in\mathbb{R}^d,\qquad \gamma\in\mathbb{R}_{\ge 0},
\qquad
n \coloneqq \|u\|_2+\varepsilon,
\qquad
s \coloneqq \frac{u}{n}\in\mathbb{S}^{d-1},
\qquad
w \coloneqq \gamma\,s.
\label{eq:explicit_param}
\end{equation}
Now, instead of directly parameterizing $w$, it becomes \emph{implicitly} defined through parameters $(u,\gamma)$.

\vspace{-0.3cm}
\paragraph{Factored (spherical) STE}

Since with a spherical parameterization there is no need to form $w$ in the forward pass, it becomes very natural to also consider placing the STE \emph{separately} on the shape and gain quantizers:
\begin{equation}
\frac{\partial \hat s}{\partial s}\approx I,
\qquad
\frac{\partial \hat\gamma}{\partial \gamma}\approx 1,
\label{eq:factored_ste}
\end{equation}
while differentiating exactly through the smooth maps $u\mapsto s$ and $(s,\gamma)\mapsto w$. Since $\hat w=\hat\gamma\hat s$, the intermediate surrogate gradients are
\begin{equation}
\frac{\partial \mathcal{L}}{\partial s}
\approx \hat\gamma\,g,
\qquad
\frac{\partial \mathcal{L}}{\partial \gamma}
\approx \hat s^\top g.
\label{eq:optB_ds_dgamma}
\end{equation}
Using the exact Jacobian of normalization $s=u/n$,
\begin{equation}
\frac{\partial s}{\partial u}
=
\frac{1}{n}\left(I-s s^\top\right)
\;\eqqcolon\;
\frac{1}{n}P_{s},
\qquad
P_{s}=I-s s^\top,
\label{eq:jac_norm_B}
\end{equation}
we obtain the surrogate gradient with respect to $u$:
\begin{equation}
\frac{\partial \mathcal{L}}{\partial u}
\approx
\frac{\hat\gamma}{n}\,P_{s}\,g,
\qquad n=\|u\|_2+\varepsilon.
\label{eq:optB_du}
\end{equation}

Combining \eqref{eq:optB_ds_dgamma} and \eqref{eq:optB_du}, one step of plain SGD (no momentum) with learning rate $\eta$ gives
\begin{equation}
u^{+}
=
u
-
\eta\,\frac{\hat\gamma}{\|u\|_2+\varepsilon}\,P_{s}\,g,
\qquad
\gamma^{+}
=
\gamma
-
\eta\,(\hat s^\top g),
\qquad
w^{+}
\;\coloneqq\;
\gamma^{+}\,\frac{u^{+}}{\|u^{+}\|_2+\varepsilon}.
\label{eq:optB_update_full}
\end{equation}
Note that our main analysis in (\ref{eq:optB_ds_dgamma})--(\ref{eq:optB_du}) holds for arbitrary optimizers, including AdamW, which is used in the experiments presented below.

\vspace{-0.3cm}
\paragraph{Geometric interpretation of \method.}
\label{sec:interpretation_optB}

\optionB (\method) optimizes explicit shape-gain variables $(u,\gamma)$ while the forward pass uses the quantized reconstruction $\hat w=\hat\gamma\hat s$. The induced one-step update for plain SGD is given in \eqref{eq:optB_update_full}; we interpret this update through its geometry.

The shape update can be viewed as an optimization step on the sphere. In particular, the Jacobian of normalization yields the projector
$P_s = I - ss^\top$, where $s=u/(\|u\|_2+\varepsilon)$ is the current normalized (unquantized) direction. Since
$\partial \mathcal{L}/\partial s \approx \hat\gamma\,g$, the update direction on $u$ is proportional to
$P_s(\partial \mathcal{L}/\partial s)$, i.e., the Riemannian gradient on $\mathbb{S}^{d-1}$ (up to the scalar factor $(\|u\|_2+\varepsilon)^{-1}$).
After the ambient step in $u$, the renormalization $s^{+}=u^{+}/(\|u^{+}\|_2+\varepsilon)$ maps back to $\mathbb{S}^{d-1}$ and acts as a retraction.
Thus, \method induces the standard ``tangent step + retraction'' update on directions, coupled with a Euclidean step on the gain $\gamma$.
This interpretation is exact for the one-step SGD analysis and remains a useful description of the induced direction update under common optimizers.

\vspace{-0.3cm}
\paragraph{Alternative parameterizations and STE assumptions.}
While Secs.~\ref{sec:optA}--\ref{sec:optB} focus on the two primary regimes, the Euclidean shadow weights with a joint STE (\optionA) and a shape-gain parameterization with factored STE (\optionB, or \method), other parameterization and STE choices are possible. In App.~\ref{sec:optBprime}, we isolate the effect of the \emph{parameterization} by using spherical shadow variables $(u,\gamma)$ but retaining a \emph{joint} STE; this yields a tangential--radial update similar in form to \method, but driven by the unquantized $(s,\gamma)$ rather than the quantized forward quantities $(\hat s,\hat\gamma)$. Conversely, in App.~\ref{sec:optE_factored} we isolate the effect of the \emph{STE placement} by keeping Euclidean shadow weights $w$ while applying a \emph{factored} (shape-gain) STE, which induces a direction--magnitude decomposition of the Euclidean update without an explicit retraction step. Finally, App.~\ref{sec:appendix_unnormalized_trick} shows an implementation trick that recovers the exact \optionA update while storing parameters in a factored form. Together, these variants clarify that both the parameterization and the STE factorization materially shape the effective optimization dynamics.
\vspace{-0.3cm}
\subsection{Practical quantizers}
\label{sec:practical}

We instantiate the shape quantizer with $d$$=$$8$ using normalized $E_8$-derived codebooks
$\mathcal{C}=\{c_1,\ldots,c_K\}\subset\mathbb{S}^{7}$.
For a normalized shape $s$, shape quantization uses exact nearest-neighbor assignment on the sphere,
\begin{equation}
Q_{\text{shape}}(s)=c_{j^\star},
\qquad
j^\star=\arg\max_{j\in\{1,\ldots,K\}}\langle s,c_j\rangle,
\label{eq:shape_nn}
\end{equation}
which is equivalent to minimizing Euclidean distance since both $s$ and $c_j$ are unit norm.
We use the normalized forms of E8P12 and our E8S codebooks.
Implementation details are given in App.~\ref{sec:e8-codebook}.

For gain quantization, we use a chi-quantile scalar quantizer.
At initialization, gains are derived from pretrained block norms, which are well approximated by a $\chi_d$ distribution under Gaussian weights~\citep{vanderouderaa2024pyramidvectorquantizationllms}.
For gain bit-width $b_{\gamma}$, we precompute $2^{b_{\gamma}}$ levels $\{\tau_\ell\}_{\ell=1}^{2^{b_{\gamma}}}$ as evenly spaced quantiles of $\chi_d$ and assign
\begin{equation}
\ell^\star=\arg\min_{\ell}|\gamma-\tau_\ell|,
\qquad
\hat\gamma=Q_{\text{gain}}(\gamma)=\tau_{\ell^\star}.
\label{eq:gain_chi}
\end{equation}
Unlike learned codebooks, these levels are specified in closed form by $(d,b_{\gamma})$ (up to numerical evaluation of the quantiles), so the gain quantizer introduces no additional learned codebook parameters and requires no extra bits-per-weight beyond the $b_{\gamma}$ bits used to index the gain level.

\subsection{Scalability and implementation}
\label{sec:impl}
\begin{wrapfigure}{r}{0.4\columnwidth}
\vspace{-1.7em}
\centering
\includegraphics[width=0.39\columnwidth]{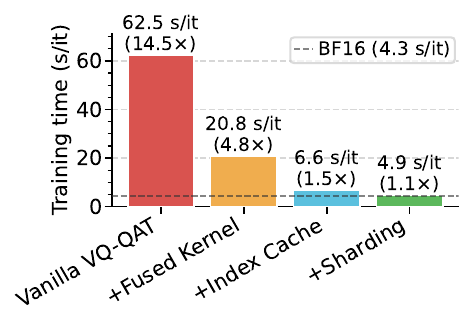}
\vspace{-1.0em}
\caption{\method reduces training time via progressively added fused kernels, index caching, and sharded assignment, approaching INT2 (4.4\,s/it) and BF16 (4.3\,s/it) performance, and allows us to scale VQ-QAT to billions of tokens.}
\vspace{-0.5em}
\label{fig:timing}
\end{wrapfigure}
Exact nearest-neighbor~(NN) search is the main computational bottleneck in VQ-QAT.
With $N$ weight blocks and a $d$-dimensional codebook of size $K \approx 2^{d\cdot \mathrm{bpw}}$, brute-force search costs $O(NKd)$ per quantization step.
Unlike PTQ, where search is performed once or a few times, QAT repeats assignment on every forward pass to simulate inference-time quantization.
In our implementation, each data-parallel worker computes exact NN~\eqref{eq:shape_nn} for a disjoint subset of vectors using a shared codebook $\mathcal{C}$, and only the resulting integer indices are synchronized via all-gather.
Furthermore, a new Triton kernel fuses normalization, codeword selection, and reconstruction.
We also cache quantized weights across gradient-accumulation steps, so assignment is invoked only once per optimizer step.
For E8S on LLaMA-3.2-1B, these optimizations reduce training time from 62.5\,s/it for vanilla VQ-QAT to 4.9\,s/it, yielding a 12.8$\times$ speedup while preserving the exact nearest-neighbor assignment, see Figure~\ref{fig:timing}.
Overall, our lattice VQ-QAT implementation incurs only moderate overhead relative to scalar QAT.

\section{Experimental setup}
\label{sec:experimental_setup}
\vspace{-0.2cm}

\paragraph{Setup.} We evaluate \method on instruction-tuned LLaMA-3.2-1B/3B-Instruct~\citep{grattafiori2024llama3herdmodels}, Qwen3-4B-Instruct-2507~\citep{qwen3technicalreport}, and base Qwen3-1.7B~\citep{qwen3technicalreport}.
We drop the ``-Instruct'' and ``-Instruct-2507'' suffixes in the text and abbreviate the model names as L1B, L3B, Q4B, and Q1B7 in tables for brevity.
QAT fine-tunes a quantized model using DCLM-Edu~\citep{allal2025smollm} dataset with Jensen--Shannon divergence (JSD) distillation~\citep{lee2025upq} from a frozen FP16 teacher.
We train for 0.25B--10B tokens with batch size 256 and context length 1024.
Learning rates are swept at 1B tokens and reused for longer runs.
All experiments use blockwise Hadamard rotations following QuIP\#~\citep{tseng2024quip-sharp-betterllmquantization}.
QAT starts with weights initialized from PTQ checkpoints using a QuIP\#-style GPTQ reconstruction with OmniQuant scale calibration.
Please find additional details in App.~\ref{app:experimental-details}.
\vspace{-0.3cm}
\paragraph{Baselines.}
Our primary scalar baseline, INT-QAT, is SEQ~\citep{liu2025paretoq,lee2025upq}, which outperformed Signed quantization in preliminary experiments (App.~\ref{app:experimental-details}).
We also compare to VQ and PTQ baselines, including QTIP and PV-Tuning where available.
\vspace{-0.3cm}
\paragraph{Evaluation metrics.}
We report WikiText2 perplexity at sequence length 4096, common-sense reasoning (CSR) and MMLU~\citep{hendrycks2021measuringmassivemultitasklanguage} accuracies.
%PIQA~\citep{bisk2020piqa}, ARC-Easy/Challenge~\citep{allenai:arc}, BoolQ~\citep{clark2019boolq}, HellaSwag~\citep{zellers2019hellaswag}, WinoGrande~\citep{ai2:winogrande}, and five-shot MMLU~\citep{hendrycks2021measuringmassivemultitasklanguage}.
The CSR score calculation follows QuIP\#~\citep{tseng2024quip-sharp-betterllmquantization} and averages over six tasks.
We also report results on long generation tasks (GSM8K~\citep{cobbe2021gsm8k}, IFEval~\citep{zhou2023ifeval}, LongBench~\citep{bai2024longbench}).
All downstream evaluations use the LLM Evaluation Harness~\citep{eval-harness}.
More details in App.~\ref{app:evaluation_protocol}.

\vspace{-0.3cm}
\section{Results}
\vspace{-0.2cm}

\paragraph{\optionA vs \optionB (\method).}
\begin{table*}[t]
\centering
\caption{Standard vs Spherical VQ-QAT (\method) results obtained with the same E8S shape-gain quantizer.
\optionA applies STE to the reconstructed block,
while \optionB applies separate STEs to the shape and gain quantizers. App.~\ref{app:ablation_standard_vs_spherical_ste_generative} extends the table to IFEval and LongBench generative tasks.
}
\label{tab:ste_ablation}
\footnotesize
\resizebox{\textwidth}{!}{%
\begin{tabular}{l ccc ccc ccc ccc}
\toprule
\multirow{2}{*}[-0.5ex]{\textbf{Method}} &
    \multicolumn{3}{c}{\textbf{WikiText2 PPL $\downarrow$}} &
    \multicolumn{3}{c}{\textbf{CSR, \% $\uparrow$}} &
    \multicolumn{3}{c}{\textbf{MMLU, \% $\uparrow$}} &
    \multicolumn{3}{c}{\textbf{GSM8K, \% $\uparrow$}} \\
\cmidrule(lr){2-4} \cmidrule(lr){5-7} \cmidrule(lr){8-10} \cmidrule(lr){11-13}
 & \textbf{L1B} & \textbf{L3B} & \textbf{Q4B}
 & \textbf{L1B} & \textbf{L3B} & \textbf{Q4B}
 & \textbf{L1B} & \textbf{L3B} & \textbf{Q4B}
 
 & \textbf{L1B} & \textbf{L3B} & \textbf{Q4B} \\
\midrule
FP16
    & 12.15 & 10.49 & 9.41
    & 61.0  & 67.5  & 72.9
    & 45.4  & 59.6  & 72.6
    & 36.8  & 65.8  & 85.4 \\
\midrule
\multicolumn{13}{l}{\textit{Tok = 1B}} \\[0.3ex]
\optionAs
    & 14.22          & 11.04          & 10.17
    & 58.4           & \textbf{66.3}  & 70.2
    & 39.3           & 55.0           & 67.1
    & 14.7           & 47.0           & 76.3 \\
\optionBs
    & \textbf{13.85} & \textbf{10.96} & \textbf{10.03}
    & \textbf{58.8}  & 65.8           & \textbf{71.0}
    & \textbf{40.3}  & \textbf{55.8}  & \textbf{67.4}
    & \textbf{18.0}  & \textbf{51.9}  & \textbf{77.0} \\
\midrule
\multicolumn{13}{l}{\textit{Tok = 5B}} \\[0.3ex]
\optionAs
    & 13.77          & 10.88          & 10.17
    & 59.1           & \textbf{66.8}  & \textbf{71.4}
    & 41.2           & \textbf{55.8}  & \textbf{68.4}
    & 15.1           & 51.8           & 78.4 \\
\optionBs
    & \textbf{13.64} & \textbf{10.84} & \textbf{10.06}
    & \textbf{59.4}  & 66.4           & 71.0
    & \textbf{41.4}  & 55.7           & \textbf{68.4}
    & \textbf{15.7}  & \textbf{52.0}  & \textbf{79.6} \\
\bottomrule
\end{tabular}
}
\end{table*}

Table~\ref{tab:ste_ablation} isolates the effect of the QAT training rule.  We compare two variants using the same E8S shape-gain quantizer and the same PTQ initialization: \optionA applies a standard STE directly to the reconstructed weight block, while \optionB applies separate STEs to the quantized direction and gain. 
This highlights the importance of a spherical parameterization together with separate shape and gain STEs.
Spherical VQ-QAT clearly improves WikiText2 perplexity and GSM8K accuracy across all tested models and training budgets, with GSM8K gains reaching up to 4.9\,\%.
CSR and MMLU improve mainly at the 1B token budget (except LLaMA-3.2-3B CSR).
% Overall, the factored shape-gain gives the clearest benefit on WikiText2 perplexity, the metric aligned with the QAT objective, while downstream metrics generally follow the same trend.
Refer to App.~\ref{app:ablation_standard_vs_spherical_ste_generative} for two more generative task results, App.~\ref{app:multiple_seeds} for per-seed evaluations, and App.~\ref{app:squat_ablation}-\ref{app:hadamard_rotations} for extra \method ablations.

\paragraph{Scaling across model family and QAT budget.}
\begin{table*}[t]
\centering
\caption{QAT scaling across instruction-tuned LLaMA and Qwen models. We report WikiText2 perplexity, CSR, and MMLU as a function of QAT token budget. Missing entries indicate results that were not run or not reported under the same protocol.
}
\label{tab:qat_scaling_models}
\footnotesize
\vspace{-.1cm}
\begin{threeparttable}
\setlength{\tabcolsep}{5.5pt}
\begin{tabular}{lcc ccc ccc ccc}
\toprule
\multirow{2}{*}[-0.5ex]{\textbf{Method}} &
\multirow{2}{*}[-0.5ex]{\textbf{Tok}} &
\multirow{2}{*}[-0.5ex]{\textbf{bpw}} &
\multicolumn{3}{c}{\textbf{WikiText2 PPL $\downarrow$}} &
\multicolumn{3}{c}{\textbf{CSR, \% $\uparrow$}} &
\multicolumn{3}{c}{\textbf{MMLU, \% $\uparrow$}} \\
\cmidrule(lr){4-6} \cmidrule(lr){7-9} \cmidrule(lr){10-12}
& & &
\textbf{L1B} & \textbf{L3B} & \textbf{Q4B} &
\textbf{L1B} & \textbf{L3B} & \textbf{Q4B} &
\textbf{L1B} & \textbf{L3B} & \textbf{Q4B} \\
\midrule
FP16 & -- & 16.0 &
12.15 & 10.49 & 9.41 &
61.0 & 67.5 & 72.9 &
45.4 & 59.6 & 72.6 \\
\midrule

QuIP\# (fine-tuned) & 52M & 2.0 &
16.67 & 11.13 & -- &
52.0 & 61.0 & -- &
26.5 & 42.1 & -- \\
QTIP & 52M & 2.0 &
14.47 & 10.12 & 8.87 &
54.3 & 62.7 & 69.5 &
33.0 & 49.4 & 62.7 \\
PV-tuning & 1B \textsuperscript{a} & 2.0 &
-- & -- & -- &
-- & -- & -- &
28.9 & 48.4 & -- \\
\midrule

INT-QAT & -- & 2.0 &
2e6 & 2e5 & 6e6 &
36.5 & 35.5 & 36.8 &
25.5 & 24.7 & 26.9 \\
INT-QAT & 1B & 2.0 &
15.07 & \textbf{10.45} & 11.13 &
57.2 & \textbf{64.8} & 68.7 &
35.3 & 52.9 & 64.4 \\
INT-QAT & 5B & 2.0 &
14.42 & \textbf{11.03} & 10.60 &
\textbf{58.7} & \textbf{66.0} & \textbf{70.5} &
39.6 & \textbf{55.4} & \textbf{67.4} \\
\midrule

\method (E8S-SG) & -- & 2.0 &
\textbf{22.12} & \textbf{18.51} & \textbf{14.78} &
\textbf{51.1} & \textbf{54.8} & \textbf{59.3} &
\textbf{27.9} & \textbf{37.5} & \textbf{46.1} \\
\method (E8S-SG) & 1B & 2.0 &
\textbf{14.76} & 11.29 & \textbf{10.50} &
\textbf{57.3} & \textbf{64.8} & \textbf{70.2} &
\textbf{36.5} & \textbf{53.1} & \textbf{64.8} \\
\method (E8S-SG) & 5B & 2.0 &
\textbf{14.33} & 11.08 & \textbf{10.41} &
57.6 & 65.9 & 69.9 &
\textbf{39.8} & 54.3 & 67.3 \\
\midrule

INT-QAT & -- & 2.25 &
2e6 & 1e5 & 2e5 &
37.0 & 35.2 & 37.1 &
23.6 & 23.0 & 23.3 \\
INT-QAT & 1B & 2.25 &
15.08 & 11.49 & 10.96 &
57.3 & 65.0 & 69.3 &
32.4 & 54.1 & 65.1 \\
INT-QAT & 5B & 2.25 &
14.42 & 11.03 & 10.57 &
58.9 & 65.8 & 70.9 &
40.2 & 54.9 & 66.9 \\
INT-QAT & 10B & 2.25 &
14.33 & 10.96 & 10.60 &
58.9 & 65.8 & \textbf{70.9} &
\textbf{41.0} & 55.5 & 67.8 \\
\midrule

\method (E8S-SG) & -- & 2.25 &
\textbf{17.90} & \textbf{15.00} & \textbf{12.52} &
\textbf{52.2} & \textbf{60.6} & \textbf{63.8} &
\textbf{29.8} & \textbf{45.8} & \textbf{54.2} \\
\method (E8S-SG) & 1B & 2.25 &
\textbf{13.85} & \textbf{10.96} & \textbf{10.03} &
\textbf{58.8} & \textbf{65.8} & \textbf{71.0} &
\textbf{40.3} & \textbf{55.8} & \textbf{67.4} \\
\method (E8S-SG) & 5B & 2.25 &
\textbf{13.64} & \textbf{10.84} & \textbf{10.06} &
\textbf{59.4} & \textbf{66.4} & \textbf{71.0} &
\textbf{41.4} & \textbf{55.7} & \textbf{68.4} \\
\method (E8S-SG) & 10B & 2.25 &
\textbf{13.74} & \textbf{10.83} & \textbf{9.90} &
\textbf{59.5} & \textbf{66.4} & 70.7 &
\textbf{41.0} & \textbf{57.3} & \textbf{69.5} \\
\bottomrule
\end{tabular}

\begin{tablenotes}[flushleft]
\footnotesize
\item[a] Performance from HuggingFace; training budget unspecified. Standard PV-tuning uses 1B RedPajama tokens~\citep{pvtuning}.
\end{tablenotes}
\end{threeparttable}
\vspace{-0.1cm}
\end{table*}

Having established the training rule, Table~\ref{tab:qat_scaling_models} evaluates how \method performs in QAT, comparing against scalar INT-QAT and strong VQ baselines.

Results show that QAT consistently improves over the corresponding PTQ initialization: at 2.25~bpw and 10B tokens, \method reduces WikiText2 perplexity from 17.90 to 13.74 on LLaMA-3.2-1B, from 15.00 to 10.83 on LLaMA-3.2-3B, and from 12.52 to 9.90 on Qwen3-4B.
Downstream gains are also substantial; for example, on Qwen3-4B, MMLU improves from 54.2 after PTQ to 69.5 after QAT. 

Compared with scalar INT-QAT, \method performs favorably overall. At 2.25~bpw and 10B training tokens, it matches or surpasses INT-QAT on nearly all evaluated models and metrics, with Qwen3-4B CSR as the only exception. At 2.0~bpw, the two methods are broadly comparable, although \method tends to perform better at lower training budgets ($<$\;5B tokens), while INT-QAT often gains an advantage with more extensive training.

Compared with QTIP, QUIP\#, and PV-tuning, \method sometimes yields higher perplexity while achieving substantially stronger downstream accuracy.

Compared with the most recent VQ-QAT method, LC-QAT~\citep{wang2026lcqat}, \method achieves higher CSR (61.2 vs.~59.8), outperforming LC-QAT by 1.4\,\% on Qwen3-1.7B after 1B training tokens.
% After 4B training tokens, the gap reduces to x\,\%.
See more details in App.~\ref{app:lc_qat}.

\vspace{-0.2cm}
\paragraph{Chi-based gain quantization.}
In App.~\ref{app:int-gain-quantizer}, we compare the chi-quantile and uniform integer gain quantizers at matched bit-widths, and include FP8 (E4M3) gains as a high-precision reference.
Results shows that a fixed chi-based scale (0 gain bits) outperforms 1-bit uniform integer gain quantization, highlighting the importance of distribution-aware gain design at extreme quantization levels.

\vspace{-0.25cm}
\paragraph{E8S vs.\ E8P12.}
\label{sec:ablation-e8-shells}

Table~\ref{tab:e8_2bit} ablates two $E_8$ codebooks on LLaMA-3.2-1B under matched 1B-token training.
As described in Sec.~\ref{sec:e8-codebooks}, the lattice can be truncated either by selecting complete spherical shells (E8S) or by using structured subsets (E8P12).
Compared with E8P12, E8S improves performance both with and without shape-gain, suggesting that shell-based codebooks better match the angular assignment used by \method.
At 2.25~bpw, \method consistently improves results for E8P12 and E8S codebooks.
% Next, we investigate the allocation of bits between the spherical codebook and the gain quantizer at a fixed total budget of 2.0 bpw.
% To this end, we vary the gain quantizer from 2 to 4 levels (0.125–0.250 bpw) adjusting the codebook bit rate accordingly.
% MMLU improves monotonically as more bits are allocated to the gain (35.51 → 36.05 → 36.49), while perplexity and CSR improvements are non-monotonic. We also observe that allocating additional bits to the gain beyond the matched-bpw baseline (1.875-bit codebook + 3-level gain, 2.073 bpw total) yields a further improvement.

\begin{table*}[t]
\centering
\footnotesize
\begin{minipage}[c]{0.55\textwidth}
\centering
\caption{The evaluation of E8S and E8P12 codebooks with (-SG) and without shape-gain on LLaMA-3.2-1B after 1B QAT.
Bits report codebook + scale/gain bits.}
\label{tab:e8_2bit}
\vspace{-.1cm}
\footnotesize
\begin{tabular}{l @{}r@{\, + \,}l@{\hspace{1em}} c c}
\toprule
\textbf{Codebook} &
\multicolumn{2}{c}{\textbf{Bits}} &
\textbf{bpw} &
\textbf{PPL / CSR / MMLU} \\
\midrule
\multirow[c]{2}{*}{E8P12}
             & 2.0   & 0.0   & 2.00 & 14.85 / 57.6 / 35.7 \\
             & 2.0   & 0.25  & 2.25 & 14.67 / 57.5 / 35.1 \\
\midrule
\multirow[c]{3}{*}{E8S}
             & 2.0   & 0.0   & 2.00 & 14.78 / 58.1 / 35.8 \\
             & 2.0   & 0.25  & 2.25 & 14.53 / 58.0 / 37.0 \\
             & 2.25  & 0.0   & 2.25 & 14.24 / 58.4 / 38.2 \\
\midrule
\makecell[c]{E8P12-SG\\(\method)} 
             & 2.0 & 0.25 & 2.25 & 13.94 / 58.7 / 38.9 \\
\midrule
\multirow[c]{2}{*}{\makecell[c]{E8S-SG\\(\method)}}
             % & 1.875 & 0.125 & 2.00 & 14.75 / 58.1 / 35.5 \\
             % & 1.802 & 0.198 & 2.00 & 14.92 / 56.2 / 36.1 \\
             & 1.75  & 0.25  & 2.00 & 14.76 / 57.3 / 36.5 \\
             % & 1.875 & 0.198 & 2.07 & 14.41 / 58.1 / 37.2 \\
             & 2.0   & 0.25  & 2.25 & 13.85 / 58.8 / 40.3 \\
\bottomrule
\end{tabular}
\vspace{-.2cm}
\end{minipage}
\hfill
\begin{minipage}[c]{0.4\textwidth}
\centering
\includegraphics[width=\columnwidth]{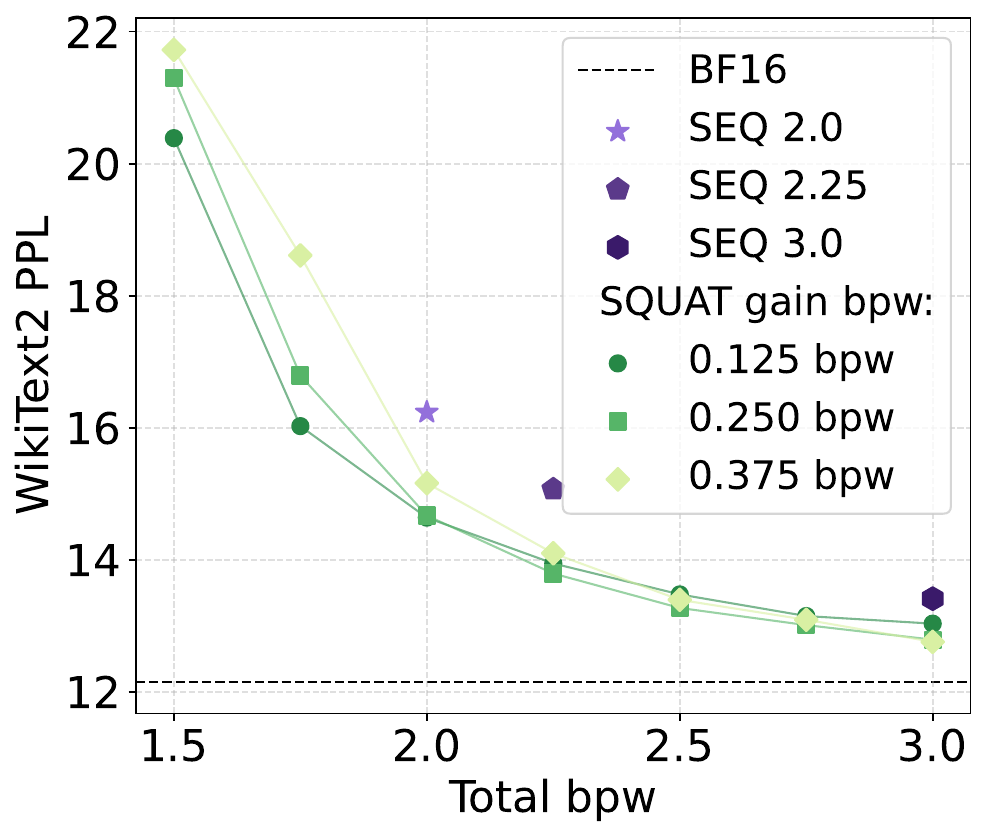}
\vspace{-0.6cm}
    \captionof{figure}{Pareto fronts of LLaMA-3.2-1B quantized with \method (E8S-SG) after 1B QAT for different gain bitwidths.}
    \label{fig:pareto_ppl_4096}
\end{minipage}
\end{table*}

\vspace{-0.25cm}
\paragraph{Pareto fronts and gain bits allocation.}
Figure~\ref{fig:pareto_ppl_4096} shows the Pareto fronts of LLaMA-3.2-1B quantized with \method (E8S-SG) after 1B-token QAT for different gain bitwidths. The optimal gain allocation depends on the target bit rate: 0.125 bpw performs best below 2.0 bpw, 0.25 bpw is preferred at up to 2.75 total bpw, while 0.375 gain bpw attains the best performance at 3.0 total bpw. Unlike INT-QAT, E8S supports fractional bitwidths, enabling finer-grained compression-accuracy trade-offs.
More detailed plots with downstream metrics can be found in App.~\ref{app:pareto_plot}.
\vspace{-0.3cm}
\section{Related work }
\vspace{-0.2cm}
\paragraph{\textbf{Post-training quantization (PTQ).}}
Post-training quantization is widely used for compressing large language models. Early approaches relied on scalar quantization with round-to-nearest strategies~\citep{pmlr-v139-hubara21a,pmlr-v119-wang20c,yao2022zeroquantefficientaffordableposttraining,dettmers2022llmint88bitmatrixmultiplication,frantar2023optimalbraincompressionframework}. GPTQ~\citep{frantar2023gptqaccurateposttrainingquantization} introduced Hessian-based reconstruction. Subsequent work addressed outliers through activation-aware scaling and sparsity~\citep{lin2024awq, dettmers2023spqrsparsequantizedrepresentationnearlossless,kim2024squeezellmdenseandsparsequantization}. 
QuIP~\citep{chee2023quip} uses Hadamard rotations to make weights more amenable to quantization, while YAQA~\citep{tseng2025modelpreservingadaptiverounding} improves Hessian estimation.

\vspace{-0.3cm}
\paragraph{\textbf{Vector and lattice quantization.}}
Vector quantization (VQ) improves low-bit rate--distortion by exploiting correlations across parameters~\citep{tseng2024quip-sharp-betterllmquantization,tseng2025qtipquantizationtrellisesincoherence, liu2024vptqextremelowbitvector, savkin2025nestquantnestedlatticequantization,vanderouderaa2026leechlatticevectorquantization}. Representative approaches include AQLM~\citep{aqlm}, which represents weights as sums of multiple codewords, and GPTVQ~\citep{vanbaalen2025gptvqblessingdimensionalityllm}, which leverages high-dimensional effects for accurate compression. However, naive VQ scales poorly due to large codebooks and expensive nearest-neighbor search.

Lattice VQ addresses this by using structured codebooks derived from dense lattices~\citep{GIBSON1988259,SpherePackings}.
QuIP\#~\citep{tseng2024quip-sharp-betterllmquantization} introduces an efficient finite subset of the $E_8$ lattice, QTIP~\citep{tseng2025qtipquantizationtrellisesincoherence} scales VQ via trellis-based decoding, and recent work explores learning lattice structure directly~\citep{zhang2025learning}.
A few recent VQ methods further decompose weights into shape and gain to allocate bits more efficiently. Pyramid Vector Quantization (PVQ)~\citep{vanderouderaa2024pyramidvectorquantizationllms} projects vectors onto the hypersphere using an implicit lattice, while PCDVQ~\citep{yue2025pcdvqenhancingvectorquantization} explicitly learns separate codebooks for shape and gain. These approaches primarily focus on post-training quantization.

\vspace{-0.3cm}
\paragraph{\textbf{Quantization-aware training (QAT).}}
Quantization-aware training uses simulated quantization and straight-through estimators during training~\citep{jacob2017quantizationtrainingneuralnetworks}. Prior work improves stability via learned clipping~\citep{choi2018pactparameterizedclippingactivation}, learned step sizes~\citep{Esser2020LEARNED}, stochastic rounding~\citep{gupta2015deeplearninglimitednumerical}, and low-rank adaptations~\citep{bondarenko2024lowrankquantizationawaretrainingllms,loeschcke2024loqtlowrankadaptersquantized}. Stretched Elastic Quantization~\citep{liu2025paretoq,lee2025upq} further improves low-bit scalar QAT, including in long-horizon LLM training.
Only limited work studies QAT with large vector or lattice codebooks.
PV-Tuning~\citep{pvtuning} introduces explicit reassignment steps, however they do not study longer token training.
LC-QAT~\citep{wang2026lcqat} enables up to 10B QAT through a linear-constrained vector-quantization parameterization.
Separately, constraining optimization to spherical manifolds has been shown to improve stability in large-scale LLM training, e.g., via spectral-norm sphere constraints~\citep{xie2026controlledllmtrainingspectral}.
DoRA~\citep{liu2024doraweightdecomposedlowrankadaptation} similarly decomposes weights into shape and gain, revealing distinct optimization dynamics in full-precision and low-rank settings.
In contrast, we study vector-quantized QAT, where applying separate STEs to shape and gain quantizers induces spherical optimization of quantized weight directions.
\vspace{-0.3cm}
\section{Discussion, future work, and limitations}
\vspace{-0.2cm}

This work studies lattice VQ-QAT at billion-parameter scale and multi-billion-token horizons.
With shape-gain parameterization and spherical $E_8$ codebooks, 
\method substantially improves over VQ-PTQ baselines and a standard joint-STE VQ-QAT baseline under the same quantizer and initialization.

\method outperforms INT-QAT at 2.25~bpw across almost all models, metrics, and QAT token budgets, while remaining competitive at 2.0~bpw. Mature INT-QAT benefits from a wide range of stabilization techniques, whereas VQ-QAT remains relatively underexplored.
%To our knowledge, this is the first study of VQ-QAT beyond 1B training tokens, enabled by an efficient distributed nearest-neighbor search implementation.
To our knowledge, this is the first demonstration of exact nearest-neighbor VQ-QAT at billion-parameter and multi-billion-token scale, enabled by an efficient distributed assignment implementation.
Future work should explore VQ-specific training improvements, including adaptations of INT-QAT techniques such as clipping.

Our evaluation is limited to E8P12 and E8S codebooks, up to 4B models, and budgets up to 10B tokens. 7B+ models, broader architectures, and higher-bit hierarchical schemes remain future work.

\citet{tseng2024quip-sharp-betterllmquantization} demonstrate the hardware efficiency of the E8P12 codebook through a custom CUDA implementation. However, E8P12 is limited to a fixed 2.0 bpw codebook rate. In contrast, E8S enables flexible rate allocation, but currently lacks an efficient dequantizer.
Recent work on Leech-lattice VQ~\citep{vanderouderaa2026leechlatticevectorquantization} suggests that fast dequantization may also be achievable for spherical lattice codebooks, making efficient E8S implementations a promising direction.
% Recent work on Leech-lattice VQ~\citep{vanderouderaa2026leechlatticevectorquantization} suggests that efficient E8S dequantization may be achievable, making it a promising direction for future work.

\vspace{-0.3cm}
\section{Conclusion}
\vspace{-0.2cm}
We presented \method, a shape-gain lattice VQ-QAT method for low-bit LLM compression.
\method decomposes each weight block into a normalized direction (shape) and scalar gain, quantizes shape with an $E_8$-derived spherical lattice codebook, quantizes gains separately, and applies separate STEs through the two quantizers.
This induces spherical optimization of quantized weight directions, and our ablations show that it improves over a standard VQ STE under the same quantizer and PTQ initialization.
To make long-horizon lattice VQ-QAT practical, we introduced distributed exact nearest-neighbor assignment for shared lattice codebooks, reducing training time from 62.5 s/it to 4.9\,s/it.
Across LLaMA-3.2-1B/3B and Qwen3-4B/1.7B models, QAT with \method substantially improves performance over VQ-PTQ baselines and a standard joint-STE VQ-QAT baseline, while remaining competitive with scalar INT-QAT at comparable bit budgets.
Our ablations further show the importance of chi-based gain quantization, gain bit rate selection, and shadow-weight updates.

\section*{AI Use Statement}
We used large language models (LLMs) to assist with implementing parts of our code and with editing and polishing the wording of the manuscript. All LLM-generated code was reviewed, tested, and verified by the authors. All experimental results, reported numbers, and tables were produced by our own experiments and entered manually. No data or claims were generated by LLMs. The authors take full responsibility for the content of this paper.

\section*{Reproducibility statement}
We make every effort to ensure that our results are reproducible. The proposed method is fully described in Sec.~\ref{sec:method} and App.~\ref{sec:e8-codebook}.
Complete experimental details, including learning rates, optimizer settings, and a training scheduler are reported in App.~\ref{app:experimental-details}.
Datasets, evaluation protocols, and data processing details are provided in Sec.~\ref{sec:experimental_setup} and App.~\ref{app:evaluation_protocol}.
We also plan to release the codebase and training scripts upon acceptance to facilitate reproducibility and support future research based on our methods.

\bibliography{iclr2027_conference}
\bibliographystyle{iclr2027_conference}

\newpage
\appendix
\fancyhead[L]{Preprint}
\onecolumn

\section{Constructing an $E_8$ codebook at a target bit budget}
\label{sec:e8-codebook}

To construct a lattice-based directional codebook with a prescribed bits-per-weight
budget $b_{\mathrm{pw}}$, we use the $E_8$ lattice as a high-dimensional spherical design.
Each codeword represents an $8$-dimensional weight block, so the bit budget determines
how many distinct directions we may store.

\paragraph{Enumerating $E_8$ lattice shells.}
The $E_8$ lattice admits the representation
\[
E_8 = 
\left\{ \, v = \tfrac{1}{2} w \;:\;
w \in \mathbb{Z}^8,\;
\begin{array}{l}
w_i \text{ all even or all odd},\\[0.2em]
\sum_i w_i \equiv 0 \pmod{4}
\end{array}
\right\}.
\]
The squared norm satisfies 
$\|v\|^2 = \tfrac{1}{4}\sum_i w_i^2$,
hence we may group all vectors of equal norm into
\[
\mathcal{S}(n^2) 
= \bigl\{\, v\in E_8 : \|v\|^2 = n^2 \bigr\},
\qquad n^2 \in \{n_{\min}^2, n_{\min}^2 + 2,\dots,n_{\max}^2\},
\]
where only even norms occur.  
Concatenating these shells in increasing norm produces a large candidate set
\[
\mathcal{C}_{\mathrm{raw}}
= 
\mathcal{S}(n_{\min}^2)
\;\cup\;
\mathcal{S}(n_{\min}^2+2)
\;\cup\;\cdots\;
\cup\;
\mathcal{S}(n_{\max}^2).
\]

\paragraph{Mapping bit budget to codebook size.}
For an $8$-dimensional block, storing an index into the codebook uses
\[
b_{\mathrm{code}}
    \;=\;
    \big\lfloor 8\, b_{\mathrm{pw}}\big\rfloor
\]
bits per vector, hence the maximal allowable number of directions is
\[
K = 2^{\,b_{\mathrm{code}}}.
\]
For the budgets used in our experiments:
\[
\begin{aligned}
b_{\mathrm{pw}} &= 1.75 
&\Rightarrow&\quad b_{\mathrm{code}}=14,\;\; K = 16{,}384,
\\[0.25em]
b_{\mathrm{pw}} &= 1.875 
&\Rightarrow&\quad b_{\mathrm{code}}=15,\;\; K = 32{,}768,
\\[0.25em]
b_{\mathrm{pw}} &= 2.0 
&\Rightarrow&\quad b_{\mathrm{code}}=16,\;\; K = 65{,}536.
\end{aligned}
\]

\paragraph{Selecting exactly $K$ codewords.}
Typically $|\mathcal{C}_{\mathrm{raw}}| \gg K$.  
We construct the final codebook $\mathcal{C}$ by:
\begin{enumerate}
\item Including full shells $\mathcal{S}(n^2)$ in order of increasing norm until adding
the next shell would exceed $K$.
\item Denoting by $\mathcal{S}(n_j^2)$ the first shell that would overflow $K$, we compute the
remainder 
\[
R \;=\; 
K - \sum_{i < j} |\mathcal{S}(n_i^2)|.
\]
From $\mathcal{S}(n_j^2)$  we choose exactly $R$ codewords via \emph{spherical
farthest-point sampling} similar to~\citep{yue2025pcdvqenhancingvectorquantization}, which greedily maximizes the minimum pairwise angular separation.
\end{enumerate}
The resulting set
\[
\mathcal{C}
=
\left(
\bigcup_{i < j} \mathcal{S}(n_i^2)
\right)
\cup
\mathrm{FPS}\bigl(\mathcal{S}(n_j^2),\,R\bigr)
\]
is our final spherical codebook $E_8$.

\paragraph{Shell selection for the $E_8$ shape--gain codebook.}
In shape--gain quantization, only the \emph{direction} of a lattice vector is retained.
Each candidate $v \in E_8$ is therefore projected onto the unit sphere,
\[
\hat{v} = \frac{v}{\|v\|}.
\]
Since $E_8$ is a lattice, it is closed under integer scaling: if $v \in E_8$, then
$2v \in E_8$. As a result, a lattice vector and any nonzero integer multiple of that vector correspond to the same normalized direction,
\[
\widehat{2v}
=
\frac{2v}{\|2v\|}
=
\frac{2v}{2\|v\|}
=
\frac{v}{\|v\|}
=
\hat{v}.
\]

For budgets 2.0 and 2.25 bpw considered in this work, this creates two redundant shell
pairs. The shells $\mathcal{S}(2)$ and $\mathcal{S}(8)$ generate identical sets
of normalized directions, and likewise for $\mathcal{S}(4)$ and
$\mathcal{S}(16)$. To avoid duplicate directions, only the larger-norm
shell from each redundant pair is used.

Applying this rule to the two shape--gain budgets:
\begin{itemize}
\item $b_{\mathrm{pw}}=2.0$: the shell pair
$\bigl(\mathcal{S}(2), \mathcal{S}(8)\bigr)$ is redundant.
Discarding $\mathcal{S}(2)$ yields the shell range
$\mathcal{S}(4), \mathcal{S}(6), \ldots, \mathcal{S}(12)$.

\item $b_{\mathrm{pw}}=2.25$: the shell pairs
$\bigl(\mathcal{S}(2), \mathcal{S}(8)\bigr)$ and
$\bigl(\mathcal{S}(4), \mathcal{S}(16)\bigr)$ are redundant.
Discarding $\mathcal{S}(2)$ and $\mathcal{S}(4)$ yields the shell range
$\mathcal{S}(6), \mathcal{S}(8), \ldots, \mathcal{S}(16)$.
\end{itemize}

In both cases, shell accumulation proceeds exactly as in the unnormalized
construction: shells are added in order of increasing squared norm until the
cumulative number of codewords reaches
$K = 2^{\lfloor 8 b_{\mathrm{pw}} \rfloor}$, with farthest-point sampling
applied to the final partial shell when necessary.

\paragraph{Redundancy in the E8P12 shape--gain codebook.}
Unlike the shell-based $E_8$ construction, the E8P12 codebook is a fixed coset subset of
$E_8 + \tfrac{1}{4}$ with $2^{16}$ explicitly enumerated vectors, so redundant directions
cannot be ruled out analytically.
Exhaustive enumeration confirms that exactly two pairs of codewords collapse to the same
unit direction after normalisation --- in both cases one vector has norm $3$ times the
other --- leaving $65{,}534$ unique directions out of $65{,}536$ total codewords.
We neglect this tiny reduction in our reported results.

\section{Alternative related choices of parameterization and STE}
While in Section~\ref{sec:method} we discussed the two most natural choices of combining STE and shape-gain quantization for QAT, other combinations of STE choices and parameterizations would also be possible. We discuss these possible combinations in this appendix and summarize them in Table~\ref{tab:ste_param_comparison}. 
\begin{table}[h]
\centering
\caption{Effect of STE choice and parameterization on the effective weight update $w^+$.}
\label{tab:ste_param_comparison}
\footnotesize
\renewcommand{\arraystretch}{1.4}
\begin{tabular}{c|c|c}
\hline
 &
\textbf{Joint / Standard STE}
&
\textbf{Factored (Shape-Gain) STE}
\\
 &
$\displaystyle
\begin{array}{c}
\frac{\partial \hat w}{\partial w} \approx I
\end{array}
$
&
$\displaystyle
\begin{array}{c}
\frac{\partial \hat s}{\partial s} \approx I , 
\frac{\partial \hat\gamma}{\partial \gamma} \approx 1
\end{array}
$
\\
\hline

\textbf{Euclidean parameterization ($w$)}
&
$\displaystyle
\begin{array}{c}
w^{+} = w - \eta\, g \\[0.4em]
\text{Sec.~\ref{sec:optA}}
\end{array}
$
&
$\displaystyle
\begin{array}{c}
w^{+}
=
w
-
\eta\!\left(
\frac{\hat\gamma}{\gamma} P_s g
+
(\hat s^\top g)\, s
\right)
\\[0.4em]
\text{Sec.~\ref{sec:optE_factored}}
\end{array}
$
\\
\hline

\textbf{Spherical parameterization ($u,\gamma$)}
&
$\displaystyle
\begin{array}{c}
u^{+}
=
u - \eta \frac{\gamma}{\|u\|} P_s g
\\
\gamma^{+}
=
\gamma - \eta (s^\top g)
\\[0.3em]
w^{+}
=
\gamma^{+}\,
\frac{u^{+}}{\|u^{+}\|}
\\[0.4em]
\text{App.~\ref{sec:optBprime}}
\end{array}
$
&
$\displaystyle
\begin{array}{c}
u^{+}
=
u - \eta \frac{\hat\gamma}{\|u\|} P_s g
\\
\gamma^{+}
=
\gamma - \eta (\hat s^\top g)
\\[0.3em]
w^{+}
=
\gamma^{+}\,
\frac{u^{+}}{\|u^{+}\|}
\\[0.4em]
\text{Sec.~\ref{sec:optB}}
\end{array}
$
\\
\hline
\end{tabular}
\end{table}

\subsection{What if we use spherical shadow weights, but keep the joint STE?}
\label{sec:optBprime}

In theory, we could consider a spherical parameterization of the shadow weights while still applying a joint STE to the overall quantizer. This choice is somewhat less natural, since the Euclidean variable $w$ is not explicitly present as a free optimization variable in the forward computation graph, yet one can still define such a surrogate for analysis. Concretely, let the optimized variables be the shape and gain parameters $(u,\gamma)$, as in Eq.~\ref{eq:explicit_param}. The forward pass constructs intermediate quantities via
\begin{align}
n \coloneqq \|u\|_2+\varepsilon, 
\qquad 
s \coloneqq \frac{u}{n} \in \mathbb{S}^{d-1}, 
\qquad 
w \coloneqq \gamma\, s,
\end{align}
followed by shape-gain quantization
\begin{align}
\hat s = Q_{\text{shape}}(s), 
\qquad 
\hat\gamma = Q_{\text{gain}}(\gamma), 
\qquad 
\hat w = \hat\gamma\, \hat s.
\end{align}
Thus, $w$ appears only as an intermediate node defined deterministically from $(u,\gamma)$, rather than as an independent parameter. Nevertheless, we can apply a joint STE at the level of the implied Euclidean mapping $w \mapsto \hat w$, treating the quantizer as identity in the backward pass,
\begin{align}
\frac{\partial \mathcal{L}}{\partial w}\approx g.
\end{align}
Backpropagation then proceeds through the smooth reparameterization $w=\gamma s(u)$. From $w=\gamma s$, we obtain
\begin{align}
\frac{\partial \mathcal{L}}{\partial \gamma}
\approx s^\top g,
\qquad
\frac{\partial \mathcal{L}}{\partial s}
\approx \gamma\, g,
\end{align}
and combining this with the normalization Jacobian
\begin{align}
\frac{\partial s}{\partial u}
=
\frac{1}{\|u\|_2+\varepsilon}(I - s s^\top)
\;\eqqcolon\;
\frac{1}{n}P_s,
\end{align}
yields
\begin{align}
\frac{\partial \mathcal{L}}{\partial u}
\approx
\frac{\gamma}{\|u\|_2+\varepsilon} \, P_s g.
\end{align}
One step of SGD with learning rate $\eta$ gives
\begin{align}
u^{+}
=
u - \eta\,\frac{\gamma}{\|u\|_2+\varepsilon} P_s g,
\qquad
\gamma^{+}
=
\gamma - \eta\,(s^\top g),
\qquad
w^{+}
\coloneqq
\gamma^{+}\,\frac{u^{+}}{\|u^{+}\|_2+\varepsilon}.
\end{align}
We conclude that just using the spherical shadow weights without factorized STE also produces a tangential update for the direction and a radial update for the gain, similar to \optionB. However, the update depends only on the unquantized quantities $(s,\gamma)$, rather than the quantized values $(\hat s,\hat\gamma)$ used in the forward pass. In contrast, \optionB replaces $(s,\gamma)$ with $(\hat s,\hat\gamma)$ in the surrogate gradients, yielding updated rules that are consistent with the quantized forward computation and therefore better aligned with the effective model seen during training.

\subsection{What if we use spherically-factored STE, but keep the Euclidean shadow weights?}
\label{sec:optE_factored}

For completeness, we also consider maintaining Euclidean shadow weights $w\in\mathbb{R}^d$ as optimized parameters while applying a shape-gain quantizer with a factored STE. This isolates the effect of the STE choice from the spherical parameterization.

In the forward pass, each vector is decomposed into gain and normalized shape
\begin{equation}
\gamma \coloneqq \|w\|_2 + \varepsilon,
\qquad
s \coloneqq \frac{w}{\gamma} \in \mathbb{S}^{d-1},
\end{equation}
followed by shape-gain quantization
\begin{equation}
\hat s = Q_{\text{shape}}(s),
\qquad
\hat\gamma = Q_{\text{gain}}(\gamma),
\qquad
\hat w = \hat\gamma\, \hat s.
\end{equation}

We apply STE separately to the component quantizers,
\begin{equation}
\frac{\partial \hat s}{\partial s}\approx I,
\qquad
\frac{\partial \hat\gamma}{\partial \gamma}\approx 1,
\end{equation}
while differentiating exactly through the smooth maps $w \mapsto (s,\gamma)$.

Let $g = \partial \mathcal{L}/\partial \hat w$. From $\hat w = \hat\gamma \hat s$, the surrogate gradients with respect to $(s,\gamma)$ are
\begin{equation}
\frac{\partial \mathcal{L}}{\partial s}
\approx \hat\gamma\, g,
\qquad
\frac{\partial \mathcal{L}}{\partial \gamma}
\approx \hat s^\top g.
\end{equation}
Using
\begin{equation}
\frac{\partial \gamma}{\partial w} = s,
\qquad
\frac{\partial s}{\partial w}
=
\frac{1}{\gamma}(I - s s^\top)
\;\eqqcolon\;
\frac{1}{\gamma} P_s,
\end{equation}
the surrogate gradient with respect to $w$ becomes
\begin{equation}
\frac{\partial \mathcal{L}}{\partial w}
\approx
\frac{\hat\gamma}{\gamma}\, P_s g
+
(\hat s^\top g)\, s.
\label{eq:optE_grad}
\end{equation}

A single SGD step with learning rate $\eta$ gives
\begin{equation}
w^{+}
=
w
-
\eta \left(
\frac{\hat\gamma}{\gamma}\, P_s g
+
(\hat s^\top g)\, s
\right).
\label{eq:optE_update}
\end{equation}

This update decomposes into a tangential component $\frac{\hat\gamma}{\gamma} P_s g$, which lies in the tangent space $T_s\mathbb{S}^{d-1}$, and a radial component $(\hat s^\top g)s$, aligned with the current direction $s$. Thus, even without an explicit spherical parameterization, the factored STE induces a direction--magnitude decomposition of the update. However, unlike \optionB, this update is applied directly in Euclidean space and is not followed by an explicit retraction onto the sphere; consequently, the optimization does not correspond to a clean Riemannian step on $\mathbb{S}^{d-1}$. Compared to \optionA, which yields $\partial \mathcal{L}/\partial w \approx g$, the factored STE introduces structure aligned with the shape-gain decomposition, but only in combination with a spherical parameterization (\optionB) does this lead to a principled tangent-step plus retraction update using quantized quantities $(\hat s,\hat\gamma)$.

\subsection{Implementation trick: placing STE such that `\optionA' updates with shape-gain parameters $(u, \gamma)$.}
\label{sec:appendix_unnormalized_trick}

Comparing `\optionA' and `\optionB' can be error prone, due to differences in parameterization, initialization, and scaling. We provide a simple implementation trick that emulates the exact same mathematical update as the Standard/Joint STE with Euclidean shadow weights, while storing the code parameters in an \emph{unnormalized} factored form $(u,\gamma)$, with an unconstrained vector $u\in\mathbb{R}^d$ and a scalar $\gamma\in\mathbb{R}^+$. The trick is to define a \emph{temporary} weight vector that scales $u$ by $\gamma$,

\begin{equation}
\tilde w \;\coloneqq\; \gamma\,u.
\label{eq:unnormalized_w}
\end{equation}
We define the forward quantizer $Q(\tilde w)$ by internally converting $\tilde w$ to shape-gain form:
\begin{equation}
\tilde\gamma \;\coloneqq\; \|\tilde w\|_2, 
\qquad
s \;\coloneqq\; \frac{\tilde w}{\tilde\gamma},
\qquad
\hat w \;\coloneqq\; Q_{\text{gain}}(\tilde\gamma)\cdot Q_{\text{shape}}(s).
\label{eq:internal_shape_gain}
\end{equation}
Crucially, we apply a \emph{joint} straight-through estimator at the outer level:
\begin{equation}
\frac{\partial \hat w}{\partial \tilde w} \;\approx\; I.
\label{eq:outer_ste}
\end{equation}
Let $g = \partial \mathcal L / \partial \hat w$ be the upstream gradient. By \eqref{eq:outer_ste}, we obtain
\begin{equation}
\frac{\partial \mathcal L}{\partial \tilde w}
\;\approx\;
g.
\label{eq:grad_tilde_w}
\end{equation}
Using $\tilde w = \gamma u$, we propagate gradients to $(u,\gamma)$:
\begin{equation}
\frac{\partial \mathcal L}{\partial u}
=
\gamma\,g,
\qquad
\frac{\partial \mathcal L}{\partial \gamma}
=
u^\top g.
\label{eq:grad_u_gamma_unnormalized}
\end{equation}

\optionA applies a joint STE directly to an unconstrained Euclidean parameter $w$, yielding
\begin{equation}
\frac{\partial \mathcal L}{\partial w} \;\approx\; g.
\label{eq:optA_app}
\end{equation}
The construction above is equivalent to \optionA under the reparameterization
\begin{equation}
w \;\equiv\; \tilde w = \gamma u.
\end{equation}
Indeed, \eqref{eq:grad_tilde_w} exactly matches \eqref{eq:optA_app}, showing that optimization proceeds as if $\tilde w$ were the Euclidean shadow weights. The gradients on $(u,\gamma)$ in \eqref{eq:grad_u_gamma_unnormalized} are simply those induced by the linear reparameterization $w = \gamma u$. Although the forward pass uses a shape-gain decomposition \eqref{eq:internal_shape_gain}, the backward pass ignores this structure due to the outer STE \eqref{eq:outer_ste}.
\section{Additional experimental details}
\label{app:experimental-details}

For PTQ initialization, we follow the QuIP\# setup~\citep{tseng2024quip-sharp-betterllmquantization}: a GPTQ-style column-wise reconstruction with Hessian-based error propagation is used to quantize pretrained weights.
For PTQ initialization and scale calibration, we follow OmniQuant~\citep{shao2023omniquant} and use 128 WikiText2 sequences of length 2048~\citep{merity2016pointer}.
Shape-gain uses gain group size 8.
For non-shape-gain models at exactly 2~bpw, we apply channel-wise post-scaling and tune Hadamard rotations.
We omit auxiliary storage overheads below 0.01~bpw from reported bit budgets; for example, one BF16 scale per channel adds approximately $16/2048\approx0.008$~bpw for hidden dimension 2048.

All scalar INT-QAT baselines use the min-error scale estimator with $p=3.0$.
We also experimented with GPTQ-style PTQ initialization for scalar baselines, but found it to match or slightly underperform SEQ after extended QAT.
The comparison of SEQ and standard Signed quantizers is shown in Table~\ref{tab:scalar_hadamard_scheduler_offset}.
\begin{table}
\centering
\caption{The influence of learning rate scheduler, Hadamard transformation, and quantizer on the performance of the quantized LLama-3.2-1B model after 1B token INT-QAT.
}
\label{tab:scalar_hadamard_scheduler_offset}
\footnotesize
\resizebox{\textwidth}{!}{%
\begin{tabular}{ccccccccc}
\toprule
\multirow{2}{*}[-0.5ex]{bpw} & \multirow{2}{*}[-0.5ex]{\textbf{Scheduler}} & \multirow{2}{*}[-0.5ex]{\textbf{Hadamard}} & \multicolumn{2}{c}{\textbf{WikiText2 perplexity $\downarrow$}} & \multicolumn{2}{c}{\textbf{CSR, \% $\uparrow$}} & \multicolumn{2}{c}{\textbf{MMLU, \% $\uparrow$}} \\
\cmidrule(lr){4-5} \cmidrule(lr){6-7} \cmidrule(lr){8-9}
&  &  & \textbf{Signed} & \textbf{SEQ} & \textbf{Signed} & \textbf{SEQ} & \textbf{Signed} & \textbf{SEQ} \\
\midrule
FP16 &  &  & \multicolumn{2}{c}{12.15} & \multicolumn{2}{c}{61.0} & \multicolumn{2}{c}{45.4} \\
\midrule
\multirow{3}{*}{2.0} & Cosine &  & 18.07 & \textbf{16.97} & 54.5 & \textbf{55.7} & 26.0 & \textbf{28.5} \\
& WSD &  & 16.91 & \textbf{16.25} & 54.7 & \textbf{56.3} & 28.1 & \textbf{30.8} \\
& WSD & \checkmark & 15.91 & \textbf{15.07} & 56.3 & \textbf{57.3} & 28.3 & \textbf{31.7} \\
\midrule
\multirow{3}{*}{2.25} & Cosine &  & 16.71 & \textbf{16.32} & 55.1 & \textbf{55.8} & \textbf{31.8} & 30.9 \\
& WSD &  & 15.96 & \textbf{15.77} & 56.1 & \textbf{56.7} & 29.5 & \textbf{32.6} \\
& WSD & \checkmark & 15.85 & \textbf{14.94} & 56.8 & \textbf{57.5} & 30.8 & \textbf{33.6} \\
% \midrule
% \multirow{3}{*}{3.0} & Cosine &  & \textbf{13.47} & 13.57 & 59.7 & \textbf{59.9} & 40.8 & \textbf{41.7} \\
% & WSD &  & \textbf{13.38} & 13.41 & 59.8 & \textbf{59.9} & \textbf{42.6} & 40.7 \\
% & WSD & \checkmark & 13.11 & \textbf{13.01} & \textbf{60.5} & \textbf{60.5} & 42.5 & \textbf{43.2} \\
% \midrule
% \multirow{3}{*}{3.25} & Cosine &  & \textbf{13.00} & 13.08 & \textbf{61.1} & 60.7 & \textbf{43.6} & 42.9 \\
% & WSD &  & \textbf{12.98} & 13.02 & \textbf{60.8} & 60.4 & \textbf{43.5} & 42.6 \\
% & WSD & \checkmark & 12.84 & \textbf{12.77} & \textbf{61.0} & \textbf{61.0} & 42.8 & \textbf{44.1} \\
\bottomrule
\end{tabular}
}
\end{table}

We use a block size of 32 and an FP8 (E4M3) gain quantizer when comparing it against the chi-quantile quantizer.
Consequently, an additional overhead of 0.25 bpw is included in the total bit rate.

QAT uses the GKD trainer with the AdamW optimizer
($\beta_1{=}0.9$, $\beta_2{=}0.999$, $\epsilon{=}10^{-8}$, weight decay~$0$).
One billion training runs for 3{,}815 steps on the DCLM-edu dataset with sequence length
1024, per-device batch size 32 across 8 GPUs (effective batch size 256,
corresponding to $\approx$1B tokens).

The learning rate schedule is \emph{warmup-stable-decay} (WSD) in all VQ and INT experiments:
10\% warmup, 80\% stable phase, 10\% cosine decay to~0.
The comparison with \emph{cosine} scheduler was done for the scalar quantizer with and without Hadamard transformation applied. Please refer to Table~\ref{tab:scalar_hadamard_scheduler_offset}.

We used a grid search to find optimal learning rates for each method except SEQ, for which we used the original learning rate from~\citep{liu2025paretoq}. 
Table~\ref{tab:lr_settings} shows the selected learning rates.
Note that for \method we always used the same learning rates to be compatible with SEQ. For \optionA Qwen3-4B, we used a slightly higher LR found during a grid search.

\begin{table}[t]
\centering
\caption{Learning rates selected for experiments.}
\label{tab:lr_settings}
\footnotesize
\resizebox{\textwidth}{!}{%
\begin{tabular}{llcccc}
\toprule
\textbf{Method}
    & \textbf{Granularity}
    & \textbf{Tokens}
    & \textbf{Weight/Shape LR}
    & \textbf{Scale/Gain LR}
    & \textbf{Model} \\
\midrule
\multirow{2}{*}{\method (E8S-SG)} & \multirow{2}{*}{all bpv variants} & \multirow{2}{*}{1B, 5B, 10B} & \multirow{2}{*}{$3 \times 10^{-5}$} & \multirow{2}{*}{$3 \times 10^{-2}$} & L1B, L3B, \\ & & & & & Q1B7, Q4B \\
\optionA
    & all bpv variants
    & 1B, 5B
    & $3 \times 10^{-5}$
    & ---
    & L1B, L3B \\
\optionA
    & all bpv variants
    & 1B, 5B
    & $1 \times 10^{-4}$
    & ---
    & Q4B \\
E8S
    & all bpv variants
    & 1B
    & $1 \times 10^{-4}$
    & $1 \times 10^{-2}$
    & L1B \\
\method (E8P12-SG)
    & all bpv variants
    & 1B
    & $3 \times 10^{-5}$
    & $3 \times 10^{-2}$
    & L1B \\
E8P12
    & all bpv variants
    & 1B
    & $1 \times 10^{-4}$
    & $1 \times 10^{-2}$
    & L1B \\
SEQ
    & all bpv variants
    & 1B, 5B, 10B
    & $2 \times 10^{-5}$
    & $2 \times 10^{-5}$
    & L1B, L3B, Q4B \\
\bottomrule
\end{tabular}
}
\end{table}

All runs use bfloat16 precision with gradient checkpointing enabled.
\section{Evaluation protocol}
\label{app:evaluation_protocol}
All evaluations use the \texttt{lm-evaluation-harness} framework~\citep{eval-harness}
with greedy decoding (\texttt{do\_sample=False}, \texttt{temperature=0}) and
batch size~1 throughout.

\subparagraph{MMLU~\citep{hendrycks2021measuringmassivemultitasklanguage}.}
Five-shot, multiple-choice log-likelihood scoring (no text generation).
The reported score is the mean accuracy across all subtasks, multiplied by 100.

\subparagraph{Common sense reasoning (CSR).}
Zero-shot, multiple-choice log-likelihood scoring across six tasks:
BoolQ~\citep{clark2019boolq}, PIQA~\citep{bisk2020piqa}, WinoGrande~\citep{ai2:winogrande}, HellaSwag~\citep{zellers2019hellaswag}, ARC-Easy/Challenge~\citep{allenai:arc}.
For each task the metric is length-normalised accuracy (\texttt{acc\_norm})
where available (ARC-Easy, ARC-Challenge, HellaSwag, PIQA),
and raw accuracy (\texttt{acc}) otherwise (BoolQ, WinoGrande).
CSR is the unweighted macro-average of these six values, multiplied by 100:
\begin{equation}
  \mathrm{CSR} = \frac{1}{6}\sum_{t \in \mathcal{T}} \mathrm{acc}^{(t)} \times 100,
  \quad
  \mathcal{T} = \{\text{BoolQ, PIQA, WinoGrande, HellaSwag, ARC-E, ARC-C}\}.
\end{equation}

\subparagraph{GSM8K~\citep{cobbe2021gsm8k}.}
Eight-shot, generative evaluation (\texttt{output\_type: generate\_until}).
Greedy decoding; generation stops at \texttt{"Question:"}, \texttt{"</s>"},
or \texttt{"<|im\_end|>"}.
Two filters are applied: a \emph{strict} filter extracting the number after
\texttt{"\#\#\#\#"}, and a \emph{flexible} filter extracting the last number
in the output; both use exact match as the metric.
\textbf{Reported metric:} \texttt{flexible-extract} exact match, multiplied by 100.
The flexible filter is preferred as it measures arithmetic correctness
independently of output formatting.

\subparagraph{IFEval.}
Zero-shot, generative evaluation on the Google IFEval benchmark~\citep{zhou2023ifeval}.
Greedy decoding; no explicit stop token; maximum 1280 new tokens.
Four metrics are available: prompt-level and instruction-level accuracy,
each in strict and loose variants.
\textbf{Reported metric:} the mean of all four accuracy variants ---
prompt-level and instruction-level, each in strict and loose form ---
multiplied by 100.
The chat template is applied during evaluation (\texttt{apply\_chat\_template=True}).

\subparagraph{LongBench-E.}
Zero-shot, generative evaluation on the English subset of LongBench-E,
covering three task groups:
Single-Document QA (\texttt{Qasper-E}, \texttt{MultifieldQA-EN-E}),
Multi-Document QA (\texttt{HotpotQA-E}, \texttt{2WikiMultihopQA-E}),
and Summarization (\texttt{GovReport-E}, \texttt{MultiNews-E}).
Greedy decoding; maximum 128 new tokens per subtask.
QA subtasks use QA-F1 score; summarisation subtasks use ROUGE.
\textbf{Reported metric:} unweighted average of the three category scores,
multiplied by 100 (following the evaluation protocol of the original
LongBench paper~\citep{bai2024longbench})).
\section{Additional experimental results}
\label{sec:ablation-shape-gain}

\subsection{Results on generative tasks}
\label{app:ablation_standard_vs_spherical_ste_generative}
\begin{table*}[t]
\centering
\caption{Ablation of the STE used in \method at 2.25~bpw across LLaMA-3.2-1B/3B and Qwen3-4B on generative tasks.
Both variants use the same E8S shape-gain quantizer and PTQ initialization.
}
\label{tab:ste_ablation_generative}
\footnotesize
\begin{tabular}{lc ccc ccc}
\toprule
\multirow{2}{*}[-0.5ex]{\textbf{Method}} &
    \multirow{2}{*}[-0.5ex]{\textbf{Tok}} &
    \multicolumn{3}{c}{\textbf{IFEval, \% $\uparrow$}} &
    \multicolumn{3}{c}{\textbf{LongBench, \% $\uparrow$}} \\
\cmidrule(lr){3-5} \cmidrule(lr){6-8}
 & & \textbf{L1B} & \textbf{L3B} & \textbf{Q4B}
   & \textbf{L1B} & \textbf{L3B} & \textbf{Q4B} \\
\midrule
FP16 & --
    & 57.0  & 73.9  & 86.3    % IFEval
    & 28.4  & 23.8  & 22.4 \\ % LongBench
\midrule
\optionA & 1B
    & 37.8           & \textbf{63.0}  & 79.8            % IFEval
    & 21.1           & 19.4           & 17.9 \\         % LongBench
\optionB\ (\method) & 1B
    & \textbf{41.4}  & 62.5           & \textbf{82.3}   % IFEval
    & \textbf{23.6}  & \textbf{21.6}  & \textbf{18.1} \\ % LongBench
\midrule
\optionA & 5B
    & 40.8           & 59.5           & 78.9            % IFEval
    & \textbf{24.5}  & 18.4           & 17.2 \\         % LongBench
\optionB\ (\method) & 5B
    & \textbf{43.6}  & \textbf{64.2}  & \textbf{82.4}   % IFEval
    & 23.4           & \textbf{20.7}  & \textbf{17.6} \\ % LongBench
\bottomrule
\end{tabular}
\vspace{-0.3cm}
\end{table*}

Table~\ref{tab:ste_ablation_generative} shows Standard vs. Spherical VQ-QAT results on two additional generative tasks, IFEval and LongBench. Across both tasks, Spherical VQ-QAT generally outperforms Standard VQ-QAT.

\subsection{Statistical stability across random seeds}
\label{app:multiple_seeds}

\begin{table}[t]
\centering
\caption{Per-seed results for Naive vs.\ Spherical gradient on LLaMA-3.2-1B-Instruct at 2.25\,bpw. PPL: WikiText2 perplexity ($\downarrow$). CSR: common-sense reasoning (\%, $\uparrow$). MMLU (\%, $\uparrow$). For 1B token training, ranges do not overlap for any metric, confirming that the Spherical improvement exceeds run-to-run noise.}
\label{tab:variance_naive_vs_spherical}
\small
\begin{tabular}{l l ccc ccc}
\toprule
& & \multicolumn{3}{c}{\textbf{Standard VQ-QAT}}
  & \multicolumn{3}{c}{\textbf{Spherical VQ-QAT (SQUAT)}} \\
\cmidrule(lr){3-5} \cmidrule(lr){6-8}
& & \textbf{PPL}
  & \textbf{CSR}
  & \textbf{MMLU}
  & \textbf{PPL}
  & \textbf{CSR}
  & \textbf{MMLU} \\
\midrule

\multicolumn{8}{l}{\textit{1B training tokens}} \\
\midrule
& Seed 0 & 14.22 & 58.43 & 39.28 & 13.85 & 58.77 & 40.26 \\
& Seed 1 & 14.29 & 58.18 & 36.54 & 13.90 & 58.68 & 39.50 \\
& Seed 2 & 14.13 & 58.27 & 38.53 & 13.78 & 58.65 & 40.73 \\
\cmidrule(lr){3-8}
& Mean   & 14.21 & 58.29 & 38.12 & 13.84 & 58.70 & 40.16 \\
& Std    & 0.08  & 0.13  & 1.42  & 0.06  & 0.06  & 0.62  \\
& Min    & 14.13 & 58.18 & 36.54 & 13.78 & 58.65 & 39.50 \\
& Max    & 14.29 & 58.43 & 39.28 & 13.90 & 58.77 & 40.73 \\

\midrule
\multicolumn{8}{l}{\textit{5B training tokens}} \\
\midrule
& Seed 0 & 13.77 & 59.08 & 41.23 & 13.64 & 59.44 & 41.39 \\
& Seed 1 & 13.96 & 59.12 & 40.33 & 13.77 & 59.27 & 40.53 \\
& Seed 2 & 13.83 & 58.72 & 40.88 & 13.71 & 59.64 & 41.69 \\
\cmidrule(lr){3-8}
& Mean   & 13.85 & 58.97 & 40.81 & 13.71 & 59.45 & 41.20 \\
& Std    & 0.10  & 0.22  & 0.45  & 0.07  & 0.19  & 0.60  \\
& Min    & 13.77 & 58.72 & 40.33 & 13.64 & 59.27 & 40.53 \\
& Max    & 13.96 & 59.12 & 41.23 & 13.77 & 59.64 & 41.69 \\

\bottomrule
\end{tabular}
\end{table}

Statistical significance over seeds is studied in Table~\ref{tab:variance_naive_vs_spherical}.
For both 1B and 5B QAT, (1) Spherical VQ-QAT (\method) outperforms Standard VQ-QAT on every metric across all three matched seeds, and (2) the differences in mean performance between Spherical and Standard VQ-QAT exceed the standard deviation (std) for both PPL and CSR.

\subsection{Explicit gain parameterization and shadow-weight updates}
\label{app:squat_ablation}
\begin{table}[t]
\centering
\caption{Ablation of explicit gain parameterization and shadow-weight updates under a fixed 2.25~bpw budget using 1B tokens and the E8S codebook.}
\label{tab:shape-gain-ablation}
\vspace{0.3em}
\footnotesize
\begin{tabular}{lc}
\toprule
\textbf{Setting} & \textbf{PPL / CSR / MMLU} \\
\midrule
\multicolumn{2}{l}{\textit{LLaMA-3.2-1B-Instruct}} \\
\midrule
Explicit gain parameter (\method, \text{Sec.~\ref{sec:optB}})
    & 13.85 / 58.8 / 40.3 \\
No separate gain parameter (\text{Sec.~\ref{sec:optE_factored}})
    & 14.01 / 58.2 / 38.4 \\
Without training shape weights
    & 13.85 / 58.2 / 40.0 \\
No shape-gain (E8S)
    & 14.24 / 58.4 / 38.2 \\
\midrule
\multicolumn{2}{l}{\textit{LLaMA-3.2-3B-Instruct}} \\
\midrule
Explicit gain parameter (\method, \text{Sec.~\ref{sec:optB}})
    & 10.96 / 65.8 / 55.8 \\
Without training shape weights
    & 11.00 / 65.8 / 55.2 \\
\midrule
\multicolumn{2}{l}{\textit{Qwen3-4B-Instruct}} \\
\midrule
Explicit gain parameter (\method, \text{Sec.~\ref{sec:optB}})
    & 10.03 / 71.0 / 67.4 \\
Without training shape weights
    & 10.08 / 71.0 / 66.7 \\
\bottomrule
\end{tabular}
\end{table}

We ablate how gain is represented and optimized under shape–gain VQ-QAT at a fixed 2.25~bpw budget.
As described in Sec.~\ref{sec:setup}, the \emph{explicit-gain} formulation used by \method introduces a learnable per-block gain parameter $\gamma_b$, which is optimized and quantized independently and applied after shape quantization.
In contrast, the \emph{implicit-gain} formulation described in Sec.~\ref{sec:optE_factored} derives the gain on-the-fly from the shadow weights as
$Q_{\text{gain}}(\|u_b\|_2)$, coupling scale and shape through the normalization.

In both cases, blocks are normalized before shape quantization.
After 1B QAT tokens (Table~\ref{tab:shape-gain-ablation}), the explicit gain achieves a lower perplexity than the implicit alternative (13.85 vs.\ 14.01) and a higher CSR and MMLU.

Freezing shape weights while training gains and Hadamard parameters degrades MMLU and CSR performance and removing shape-gain further worsens metrics. This ablation highlights the benefit of  decoupling shape and gain.

\begin{figure}[t]
\centering
\begin{subfigure}[t]{0.32\linewidth}
    \centering
    \includegraphics[width=\linewidth]{figures/scaleup/llama-3.2-1b-wiki2-perplexity.pdf}
    \caption{WikiText2 perplexity.}
    \label{fig:appendix_llama32_1b_wiki2}
\end{subfigure}
\hfill
\begin{subfigure}[t]{0.32\linewidth}
    \centering
    \includegraphics[width=\linewidth]{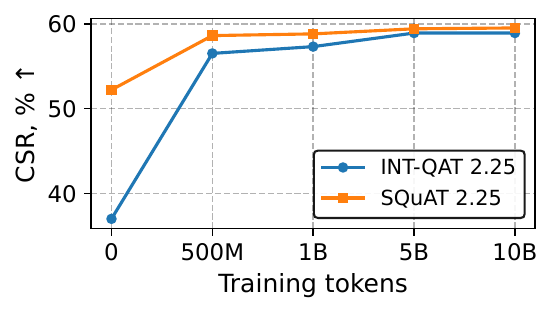}
    \caption{CSR.}
    \label{fig:appendix_llama32_1b_csr}
\end{subfigure}
\hfill
\begin{subfigure}[t]{0.32\linewidth}
    \centering
    \includegraphics[width=\linewidth]{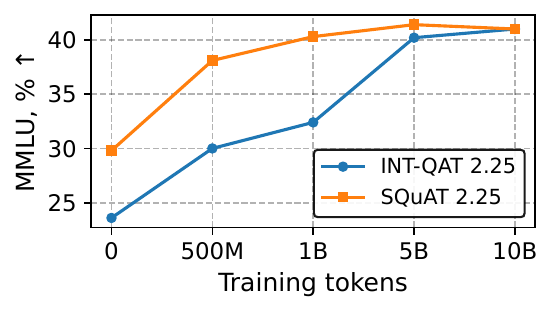}
    \caption{MMLU.}
    \label{fig:appendix_llama32_1b_mmlu}
\end{subfigure}

\caption{
Scaling behavior on LLaMA-3.2-1B as a function of QAT training tokens.
We compare INT-QAT and \method across WikiText2 perplexity, CSR, and MMLU.
}
\label{fig:appendix_llama32_1b_scaleup}
\end{figure}

\subsection{Isolating the effect of the STE choice from the spherical parameterization}
\begin{table*}[t]
\centering
\caption{Additional comparison of Euclidean parametrization and  Standard vs Spherical VQ-QAT results.
Both variants use the same E8S shape-gain quantizer and PTQ initialization.
\optionA applies STE to the reconstructed block,
while \optionB applies separate STEs to the shape and gain quantizers.
App.~\ref{app:ablation_standard_vs_spherical_ste_generative} shows results on multiple generative tasks.
}
\label{tab:ste_ablation_b2}
\footnotesize
\setlength{\tabcolsep}{3.5pt}  % default is 6pt
\begin{tabular}{lc ccc ccc ccc}
\toprule
\multirow{2}{*}[-0.5ex]{\textbf{Method}} &
    \multirow{2}{*}[-0.5ex]{\textbf{Tok}} &
    \multicolumn{3}{c}{\textbf{WikiText2 PPL $\downarrow$}} &
    \multicolumn{3}{c}{\textbf{CSR, \% $\uparrow$}} &
    \multicolumn{3}{c}{\textbf{MMLU, \% $\uparrow$}} \\
\cmidrule(lr){3-5} \cmidrule(lr){6-8} \cmidrule(lr){9-11}
 & & \textbf{L1B} & \textbf{L3B} & \textbf{Q4B}
   & \textbf{L1B} & \textbf{L3B} & \textbf{Q4B}
   & \textbf{L1B} & \textbf{L3B} & \textbf{Q4B} \\
\midrule
FP16 & --
    & 12.15 & 10.49 & 9.41    % PPL
    & 61.0  & 67.5  & 72.9    % CSR
    & 45.4  & 59.6  & 72.6 \\ % MMLU
\midrule
\optionA (Sec.~\ref{sec:optA}) & 1B
    & 14.10          & 11.04          & 10.17           % PPL
    & 58.7           & \textbf{66.3}  & 70.2            % CSR
    & 37.5           & 55.0           & 67.1 \\         % MMLU
Euclidian $w$ + factored STE (Sec.~\ref{sec:optE_factored}) & 1B
    & 14.01          & 11.11          & 10.15           % PPL
    & 58.2           & 66.0           & 70.8            % CSR
    & 38.4           & 54.6           & 66.5 \\         % MMLU
\optionB\ (\method, Sec.~\ref{sec:optB}) & 1B
    & \textbf{13.85} & \textbf{10.96} & \textbf{10.03}   % PPL
    & \textbf{58.8}  & 65.8           & \textbf{71.0}    % CSR
    & \textbf{40.3}  & \textbf{55.8}  & \textbf{67.4} \\ % MMLU
\bottomrule
\end{tabular}
\end{table*}

Table~\ref{tab:ste_ablation_b2} isolates the contribution of each component of \method:
the spherical parameterization and the factored STE.
Replacing \optionA's joint STE with a factored STE while keeping Euclidean shadow weights
(Sec.~\ref{sec:optE_factored}) yields inconsistent results---sometimes marginally better,
sometimes worse than \optionA, with no clear trend across models or metrics.
This is consistent with the analysis in Sec.~\ref{sec:optE_factored}: the factored STE in
Euclidean space does induce a tangential--radial decomposition of the gradient update,
but since the update is applied in flat Euclidean space without a retraction step,
it does not correspond to a principled Riemannian update and the induced structure
does not reliably translate into better optimization.

By contrast, \optionB\ (\method) consistently improves over \optionA\ across all three
models and all three metrics except Llama-3.2-3B CSR.
In \optionB, the factored STE is applied within the spherical parameterization,
so the shape update is automatically projected onto the tangent space and followed by
retraction via renormalization.
We therefore conclude that neither component alone is sufficient:
the spherical parameterization provides the geometric structure for a principled
tangent-step plus retraction update, while the factored STE ensures the update
is consistent with the quantized forward quantities $(\hat{s}, \hat{\gamma})$ seen during training.

\subsection{Comparison with LC-QAT}
\label{app:lc_qat}
\begin{table}[h]
\centering
\caption{Qwen3-1.7B quantization results at 2.0 total bpw. We use E8S-SG codebook in \method method. PPL is WikiText2. CSR is the average over 6 common sense reasoning tasks. LC-QAT and \method both use 1B training tokens. PPL and MMLU are not reported for the 1B-token LC-QAT. AE, AC, BQ, HS, WG, and PQ denote ARC-Easy, ARC-Challenge, BoolQ, HellaSwag, WinoGrande, and PIQA, respectively.}
\label{tab:lc_qat}
\setlength{\tabcolsep}{4pt}  % default is 6pt
\begin{tabular}{lcc cccccc c c}
\toprule
& PPL$_{2048}$$\downarrow$ & PPL$_{4096}$$\downarrow$ & AC & AE & BQ & HS & PQ & WG & CSR$\uparrow$ & MMLU$\uparrow$ \\
\midrule
FP16      & 16.72 & 15.68 & 43.08 & 69.69 & 77.52 & 60.37 & 72.14 & 61.80 & 64.10 & 60.12 \\
LC-QAT    & --    & --    & 39.51 & 64.35 & 70.43 & \textbf{55.01} & \textbf{71.65} & 57.54 & 59.75 & --    \\
\method & 17.36 & 16.19 & \textbf{39.76} & \textbf{65.87} & \textbf{75.66} & 54.06 & 69.97 & \textbf{61.72} & \textbf{61.17} & 52.94 \\
\bottomrule
\end{tabular}
\end{table}
%
% \begin{table}[h]
% \centering
% \caption{Qwen3-1.7B quantization results at 2.0 total bpw. We use E8S-SG codebook in \method method. PPL is WikiText2. CSR is the average over 6 common sense reasoning tasks. PPL at 1B-token and MMLU are not reported in LC-QAT. AE, AC, BQ, HS, WG, and PQ denote ARC-Easy, ARC-Challenge, BoolQ, HellaSwag, WinoGrande, and PIQA, respectively.}
% \label{tab:lc_qat}
% \setlength{\tabcolsep}{4pt}  % default is 6pt
% \begin{tabular}{lcc cccccc c c}
% \toprule
% & PPL$_{2048}$$\downarrow$ & PPL$_{4096}$$\downarrow$ & AC & AE & BQ & HS & PQ & WG & CSR$\uparrow$ & MMLU$\uparrow$ \\
% \midrule
% FP16      & 16.72 & 15.68 & 43.08 & 69.69 & 77.52 & 60.37 & 72.14 & 61.80 & 64.10 & 60.12 \\
% \midrule
% \multicolumn{11}{l}{\textit{1B training tokens}} \\
% \midrule
% LC-QAT    & --    & --    & 39.51 & 64.35 & 70.43 & \textbf{55.01} & \textbf{71.65} & 57.54 & 59.75 & --    \\
% \method   & 17.36 & 16.19 & \textbf{39.76} & \textbf{65.87} & \textbf{75.66} & 54.06 & 69.97 & \textbf{61.72} & \textbf{61.17} & 52.94 \\
% \midrule
% \multicolumn{11}{l}{\textit{4B training tokens}} \\
% \midrule
% LC-QAT    & 13.44 & -- & 41.72 & 68.18 & 69.85 & 58.23 & 72.69 & 57.93 & 61.43 & --    \\
% \method   & 17.21 & 16.08 & 39.42 & 66.46 & 76.06 & 55.45 & 70.18 & 60.06 & 61.27 & -- \\
% \bottomrule
% \end{tabular}
% \end{table}
Table~\ref{tab:lc_qat} compares \method against LC-QAT on Qwen3-1.7B. Since the original LC-QAT paper reports only the six CSR task accuracies for its 1B-token checkpoint, PPL and MMLU values are unavailable for direct comparison. Nevertheless,  after 1B training tokens, \method achieves higher CSR (61.2 vs.~59.8), outperforming LC-QAT by 1.4\,\%.
% After 4B training tokens, the gap reduces to x\,\%.
These results demonstrate that exact codebook-based VQ-QAT can match or exceed the performance of recent scalable VQ-QAT approaches despite directly optimizing discrete codeword assignments.

\subsection{Hadamard rotations}
\label{app:hadamard_rotations}
% \begin{minipage}[t]{0.47\textwidth}
% \centering
% \caption{Ablation of Hadamard rotations with and without shape-gain on LLaMA-3.2-1B after 1B QAT.}
% \label{tab:hadamard-ablation}
% \vspace{-.1cm}
% \footnotesize
% \begin{tabular}{lccc}
% \toprule
% \textbf{Setting} & \textbf{bpw} & \textbf{Hadamard} & \textbf{PPL / CSR / MMLU} \\
% \midrule
% \multirow{3}{*}{\makecell[l]{\method \\ (grouped)}} 
%     & \multirow{3}{*}{2.25} & Yes & 13.85 / 58.8 / 40.3 \\
%     &                       & No  & 14.89 / 58.2 / 36.5 \\
%     &                       & $\Delta$ & 1.04 / 0.6 / 3.8 \\
% \midrule
% \multirow{3}{*}{\makecell[l]{No SG \\ (grouped)}}
%     & \multirow{3}{*}{2.25} & Yes & 14.53 / 58.0 / 37.0 \\
%     &                       & No  & 15.26 / 57.1 / 33.2 \\
%     &                       & $\Delta$ & 0.73 / 0.9 / 3.8 \\
% \midrule
% \multirow{3}{*}{\makecell[l]{No SG \\ (channel)}}
%     & \multirow{3}{*}{2.25} & Yes & 14.24 / 58.4 / 38.2 \\
%     &                       & No  & 15.89 / 57.7 / 35.0 \\
%     &                       & $\Delta$ & 1.65 / 0.7 / 3.2 \\
% \midrule
% \multirow{3}{*}{\makecell[l]{No SG \\ (channel)}}
%     & \multirow{3}{*}{2.00} & Yes & 14.78 / 58.1 / 35.8 \\
%     &                       & No  & 15.53 / 57.5 / 33.3 \\
%     &                       & $\Delta$ & 0.75 / 0.6 / 2.5 \\
% \bottomrule
% \end{tabular}
% \end{minipage}
%
\begin{table}[t]
\centering
\caption{Ablation of Hadamard rotations with and without shape-gain on LLaMA-3.2-1B (E8S-SG) after 1B QAT.}
\label{tab:hadamard-ablation}
\vspace{-.1cm}
\footnotesize
\begin{tabular}{lccc}
\toprule
\textbf{Setting} & \textbf{bpw} & \textbf{Hadamard} & \textbf{PPL / CSR / MMLU} \\
\midrule
\multirow{3}{*}{\makecell[l]{\method \\ (grouped)}}
    & \multirow{3}{*}{2.25} & Yes & 13.85 / 58.8 / 40.3 \\
    &                       & No  & 14.89 / 58.2 / 36.5 \\
    &                       & $\Delta$ & 1.04 / 0.6 / 3.8 \\
\midrule
\multirow{3}{*}{\makecell[l]{No SG \\ (grouped)}}
    & \multirow{3}{*}{2.25} & Yes & 14.53 / 58.0 / 37.0 \\
    &                       & No  & 15.26 / 57.1 / 33.2 \\
    &                       & $\Delta$ & 0.73 / 0.9 / 3.8 \\
\midrule
\multirow{3}{*}{\makecell[l]{No SG \\ (channel)}}
    & \multirow{3}{*}{2.25} & Yes & 14.24 / 58.4 / 38.2 \\
    &                       & No  & 15.89 / 57.7 / 35.0 \\
    &                       & $\Delta$ & 1.65 / 0.7 / 3.2 \\
\midrule
\multirow{3}{*}{\makecell[l]{No SG \\ (channel)}}
    & \multirow{3}{*}{2.00} & Yes & 14.78 / 58.1 / 35.8 \\
    &                       & No  & 15.53 / 57.5 / 33.3 \\
    &                       & $\Delta$ & 0.75 / 0.6 / 2.5 \\
\bottomrule
\end{tabular}
\end{table}

\begin{table}[t]
\centering
\caption{Ablation of Hadamard rotations with and without shape-gain on LLaMA-3.2-1B (E8P12-SG) after 1B QAT.}
\label{tab:hadamard-ablation-e8p12}
\vspace{-.1cm}
\footnotesize
\begin{tabular}{lccc}
\toprule
\textbf{Setting} & \textbf{bpw} & \textbf{Hadamard} & \textbf{PPL / CSR / MMLU} \\
\midrule
\multirow{3}{*}{\method}
    & \multirow{2}{*}{2.25} & Yes & 13.94 / 58.7 / 38.9 \\
    &                       & No  & 14.36 / 58.3 / 36.9 \\
    &                       & $\Delta$ & 0.42 / 0.3 / 1.9 \\
\midrule
\multirow{3}{*}{No SG}
    & \multirow{2}{*}{2.25} & Yes & 14.67 / 57.5 / 35.1 \\
    &                       & No  & 15.44 / 57.2 / 33.8 \\
    &                       & $\Delta$ & 0.77 / 0.3 / 1.3 \\
\midrule
\multirow{3}{*}{\makecell[l]{No SG \\ (channel)}}
    & \multirow{2}{*}{2.00} & Yes & 14.85 / 57.6 / 35.7 \\
    &                       & No  & 15.69 / 56.9 / 32.5 \\
    &                       & $\Delta$ & 0.84 / 0.7 / 3.3 \\
\bottomrule
\end{tabular}
\end{table}

Tables~\ref{tab:hadamard-ablation} and~\ref{tab:hadamard-ablation-e8p12} study the effect of blockwise Hadamard rotations after 1B-token QAT on LLaMA-3.2-1B. Across all configurations, Hadamard rotations consistently improve performance. Removing rotations increases perplexity by 0.42--1.65 points and decreases MMLU by 1.9--3.8 points, indicating that input mixing remains beneficial even after QAT.

At the same time, shape-gain provides substantial gains both with and without Hadamard rotations. For example, in the E8S-based setting at 2.25~bpw, \method improves perplexity from 14.53 to 13.85 and MMLU from 37.0 to 40.3 when rotations are enabled. Importantly, the advantage persists even when rotations are removed, where \method achieves 14.89 perplexity and 36.5 MMLU compared to 15.26 and 33.2 without shape-gain. Similar trends are observed for E8P12-based \method. These results suggest that Hadamard rotations and shape-gain offer complementary benefits: Hadamard improves quantization by equalizing statistics, while shape-gain consistently improves accuracy regardless of whether rotations are present.

While Hadamard rotations are beneficial, unfused rotations introduce additional latency overhead and may be unsupported on some deployment targets.
Nevertheless, shape-gain retains most of its benefit even without Hadamard rotations, making E8P12-based \method particularly attractive for deployment.

\subsection{Pareto plot of E8S-based \method vs. SEQ.}
\label{app:pareto_plot}
\begin{figure}[t]
  \centering
  \includegraphics[width=\linewidth]{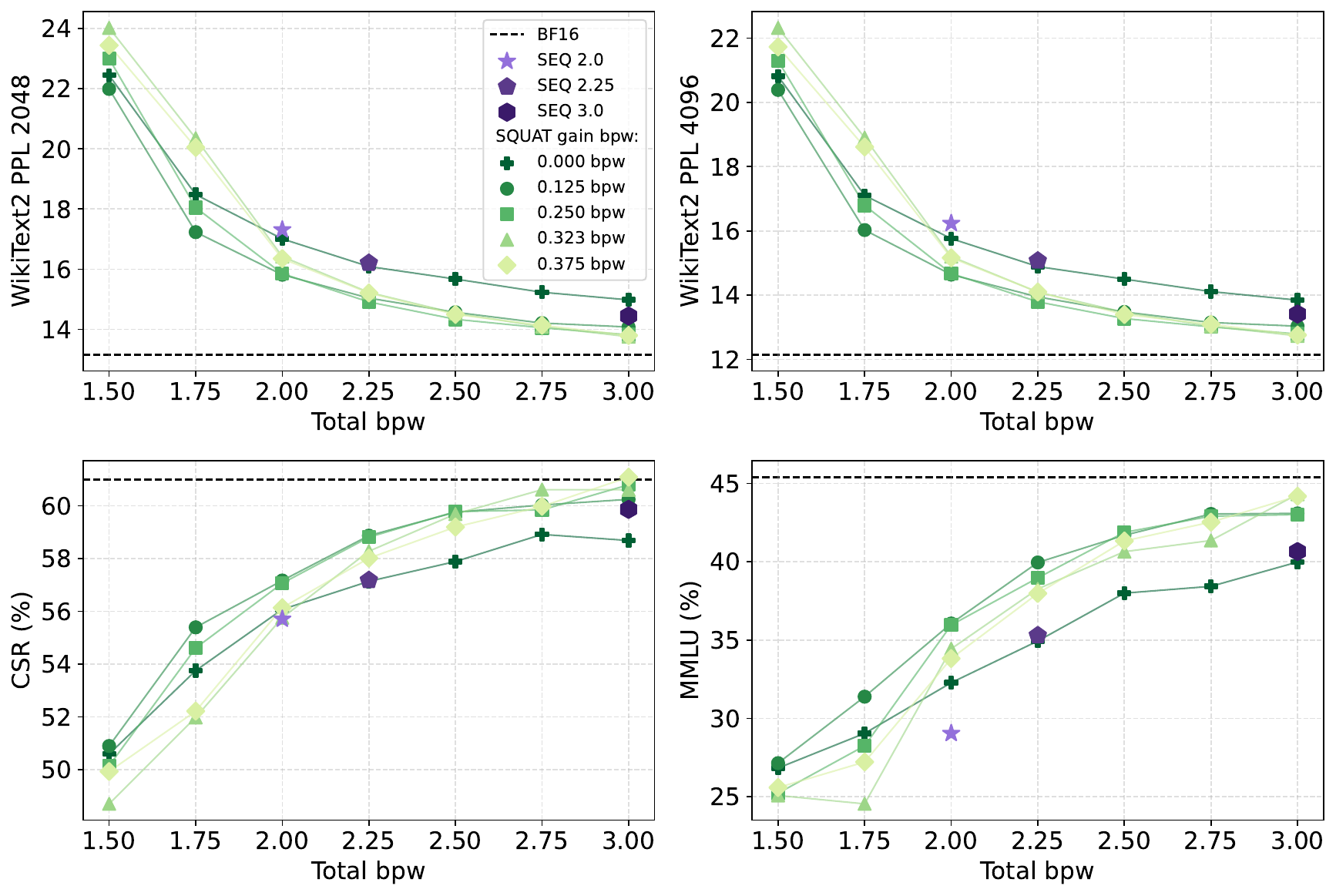}
  \caption{Pareto fronts of LLaMA-3.2-1B quantized with \method (E8S-SG) vs.\ SEQ after 1B-token QAT. Each frontier corresponds to a different gain bitwidth. The normalized E8S spherical codebook supports fractional bitwidths at finer granularity than scalar quantization, enabling more precise rate-distortion trade-offs.}
  \label{fig:pareto}
\end{figure}
Figure~\ref{fig:pareto} shows Pareto plots for a range of total bpw values and various gain quantization rates. For total bitwidth below 2.0~bpw, a gain of 0.125~bpw is preferred, while for higher total bpw, the quantization of the gain matters less.
Furthermore, our proposed E8S normalized spherical codebook construction enables fractional bitwidth at finer granularity than scalar quantization.

\section{Chi-based vs grouped integer gain quantization}
\label{app:int-gain-quantizer}

We consider a grouped integer gain quantizer as an alternative to chi-based gain quantizer.
Gains $\{\gamma_b\}_{b=1}^B$ are partitioned into groups of size $G$ (e.g., $G=128$).
For each group $t$, we compute an absmax scale
\begin{equation}
a_t \;=\; \max_{b \in \mathcal{I}_t} \gamma_b,
\qquad
s_t \;=\; \frac{a_t}{2^{b_{\gamma}}-1},
\end{equation}
and quantize each gain as
\begin{equation}
\hat \gamma
\;=\;
s_t \cdot
\mathrm{clip}\!\left(
\left\lfloor \frac{\gamma}{s_t} \right\rceil,\,
0,\,
2^{b_{\gamma}}-1
\right),
\qquad b \in \mathcal{I}_t .
\end{equation}

\begin{figure}[t]
\centering
\includegraphics[width=0.5\columnwidth]{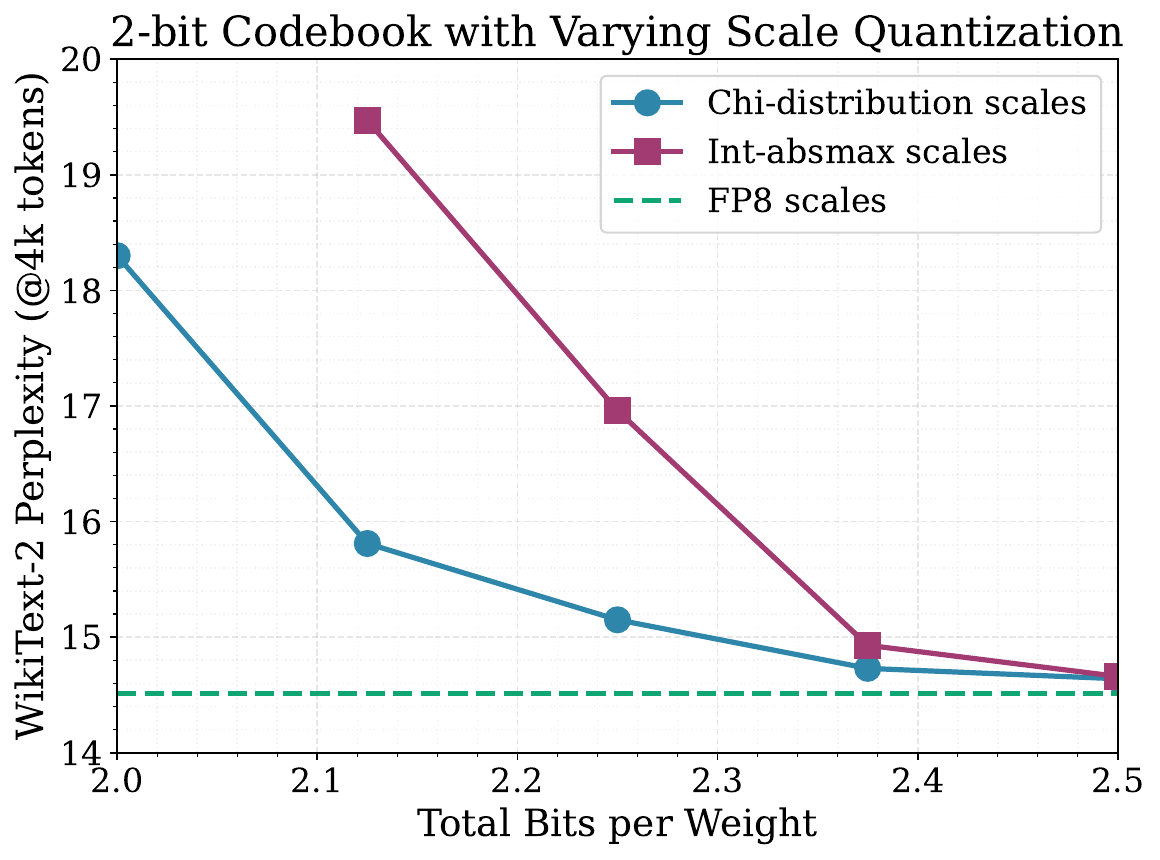}
\captionof{figure}{Scale-quantization ablation with shape-gain. Chi-based gains perform best at very low bit-rates, while all methods approach FP8-level quality as gain precision increases.}
\label{fig:qscale_ablation}
\vspace{-0.6cm}
\end{figure}

This quantizer is distribution-agnostic and simple to implement,
but we find it less effective than the chi-quantile quantizer
at very low bit-widths, particularly when $b_{\gamma} \le 2$.
Figure~\ref{fig:qscale_ablation} shows results for a 2-bit E8P12 codebook.
At very low precision (1–2 gain bits, corresponding to $\sim 0.125 - 0.25$ additional bpw), chi-quantile gains consistently outperform uniform integer quantization.
As gain precision increases, the gap narrows and both methods approach FP8 performance.
Notably, a fixed chi-based scale (zero gain bits) outperforms 1-bit uniform integer gain quantization, highlighting the importance of distribution-aware gain design in the extremely low-bit regime.
We therefore adopt chi-quantile gain quantization as the default for low-bit \method.

% This is consistent with the gain-quantization ablation in
% Figure~\ref{fig:qscale_ablation}, where chi-based gains outperform
% integer gains at low gain precision.

% We ablate the gain quantizer used in shape-gain VQ-QAT with 250M training tokens.
% We compare the chi-quantile quantizer against uniform integer gain quantization at matched bit-widths, and include FP8 (E4M3) gains as a high-precision reference.

% Figure~\ref{fig:qscale_ablation} shows results for a 2-bit E8P12 codebook.
% At very low precision (1–2 gain bits, corresponding to $\sim$0.125–0.25 additional bpw), chi-quantile gains consistently outperform uniform integer quantization.
% As gain precision increases, the gap narrows and both methods approach FP8 performance.
% Notably, a fixed chi-based scale (zero gain bits) outperforms 1-bit uniform integer gain quantization, highlighting the importance of distribution-aware gain design in the extremely low-bit regime.
% We therefore adopt chi-quantile gain quantization as the default for low-bit \method.

\end{document}